\documentclass{article}
\usepackage[preprint]{neurips_2026}
\usepackage{neurips_2026}

\usepackage[utf8]{inputenc} % allow utf-8 input
\usepackage[T1]{fontenc}    % use 8-bit T1 fonts
\usepackage{hyperref}       % hyperlinks
\usepackage{url}            % simple URL typesetting
\usepackage{booktabs}       % professional-quality tables
\usepackage{amsfonts}       % blackboard math symbols
\usepackage{amsmath}
\usepackage{nicefrac}       % compact symbols for 1/2, etc.
\usepackage{microtype}      % microtypography
\usepackage{xcolor}         % colors
\usepackage{graphicx}
\usepackage{subcaption}
\usepackage{lipsum}
\usepackage{algorithm}
\usepackage{algorithmic}
\usepackage{booktabs}
\usepackage{arydshln}
\usepackage{xcolor,tikz}

\newtheorem{thm}{Theorem}

\title{Model-Driven Policy Optimization in Differentiable Simulators via Stochastic Exploration}

\author{%
  Yuval Aroosh, Ayal Taitler \\
  Department of Industrial Engineering and Management\\
  Ben-Gurion University of the Negev\\
  Be'er Sheva, Israel \\
  \texttt{yuvalaro@post.bgu.ac.il, ataitler@bgu.ac.il} \\
}

\begin{document}

\maketitle

\begin{abstract}
Differentiable planning enables gradient-based optimization of decision-making problems by leveraging differentiable models of system dynamics. However, in highly nonlinear and hybrid discrete–continuous domains, the resulting optimization landscapes are often ill-conditioned, with flat regions and sharp transitions that hinder effective optimization.
We propose Model-Driven Policy Optimization (MDPO), a framework that introduces stochastic exploration into differentiable planning by injecting noise into the action space during optimization. Leveraging access to the model, MDPO further adapts the noise magnitude based on gradient-derived sensitivity of the trajectory objective, yielding a time-dependent exploration profile. This enables improved exploration of the objective landscape and helps escape poor local optima via dynamic allocation of exploration across timesteps and iterations.
Experiments on benchmark domains demonstrate that MDPO consistently outperforms deterministic differentiable planning, including both the noise-free variant of our method and available state-of-the-art implementations, as well as model-free baselines such as PPO, significantly improving solution quality across challenging nonlinear and hybrid settings. We further analyze the evolution of the adaptive noise magnitude across both time steps and optimization iterations, providing insight into how exploration is allocated during learning.
\end{abstract}

\section{Introduction}

Many real-world decision-making systems operate in environments where structured models of the dynamics are available or can be constructed. From aerospace and robotics to energy systems and logistics, control pipelines routinely rely on models encoding physical laws and domain knowledge. Their success highlights the effectiveness of model-based decision making.

At the same time, model-free learning, particularly reinforcement learning (RL) \cite{sutton1998reinforcement}, has achieved strong results in domains where accurate models are unavailable. These methods rely on interaction with the environment and incorporate explicit exploration mechanisms, such as stochastic policies or action perturbations, to discover high-quality solutions \cite{williams1992simple,haarnoja2018soft,lillicrap2016continuous, pathak2017curiosity}. While effective, they do not explicitly exploit known system structure. When models are available, a key question is how to leverage them for improved decision quality, sample efficiency, and robustness.

This has led to \textit{differentiable planning} \cite{schulman2015gradient}, which formulates planning as a differentiable computation graph and enables gradient-based optimization of actions or policies via backpropagation. Planning by backpropagation \cite{wu2017scalable} demonstrated this approach in deterministic settings, later extended to stochastic domains via reparameterization \cite{bueno2019deep} and to hybrid discrete–continuous systems using differentiable relaxations \cite{gimelfarb2024jaxplan}. These methods provide a unified framework for model-based optimization through automatic differentiation.

However, extending differentiable planning to complex domains typically requires smooth relaxations of discrete or logical structure. While these relaxations enable gradient propagation, they can distort the underlying problem and introduce additional plateaus and sharp transitions in the optimization landscape \cite{taitler2024ipc}. More broadly, differentiable planning lacks an explicit mechanism for exploration, relying instead on deterministic gradient updates that may be insufficient in complex landscapes.

As a result, differentiable planning often struggles in highly nonlinear and hybrid domains due to ill-conditioned and highly non-convex optimization landscapes. Gradients become uninformative in flat or saturated regions \cite{pascanu2013difficulty}, while relaxations exacerbate these effects. Optimization is therefore dominated by poor local minima, slow convergence, and sensitivity to initialization and hyperparameters. Crucially, these challenges persist even when the underlying model is accurate, suggesting that the primary limitation lies not only in modeling, but in the \textit{optimization dynamics} used to solve the resulting problems.

% We introduce \textit{Model-Driven Policy Optimization} (MDPO), a framework that improves optimization in differentiable planning via stochastic exploration. We inject noise into the action space during optimization, enabling exploration of the objective landscape and escape from poor local optima. Unlike standard RL exploration, the noise is applied within the planning model and can be adaptively scaled using information derived from the computation graph, enabling structure-aware, model-driven exploration. We show that this significantly improves performance in challenging nonlinear and hybrid domains without modifying the underlying model or its relaxation.

We introduce \textit{Model-Driven Policy Optimization} (MDPO), a framework that improves optimization in differentiable planning via stochastic exploration. We inject noise into the action space during optimization, drawing inspiration from stochastic gradient methods and variance reduction techniques \cite{greensmith2004variance}, enabling exploration of the objective landscape and escape from poor local optima. Unlike standard RL exploration, the noise is applied within the planning model and can be adaptively scaled using information derived from the computation graph, enabling structure-aware, model-driven exploration. Unlike standard stochastic optimization or gradient noise methods, MDPO introduces structured, model-driven perturbations at the level of action trajectories, allowing exploration that respects system dynamics and targets regions of high objective sensitivity.

\textbf{Contributions.}
(i) We identify optimization pathologies in differentiable planning, particularly in hybrid domains;
(ii) we propose a stochastic action-space exploration mechanism with adaptive, model-driven noise scaling;
(iii) we demonstrate substantial improvements in solution quality and convergence across benchmark problems.

\section{Background and Optimization Challenges in Differentiable Planning}

\paragraph{Markov decision processes.}
We consider a finite-horizon Markov decision process (MDP) with state space $\mathcal{S}$, action space $\mathcal{A}$, transition model $p(s_{t+1} | s_t, a_t)$, and reward function $r(s_t, a_t)$. Starting from an initial state $s_0$, the objective is to maximize the expected cumulative reward over a horizon $H$, by finding a policy $\pi:\mathcal{S}\rightarrow\mathcal{A}$:
\begin{equation}
    J(\pi) = \mathbb{E}\left[\sum_{t=0}^{H-1} r(s_t, a_t)\right].
    \label{eq:mdp_objective}
\end{equation}
The expectation is taken w.r.t. the trajectory distribution induced by $a_t=\pi(s_t)$ and $p(s_{t+1} | s_t, a_t)$.

\subsection{Differentiable Planning in Continuous, Stochastic, and Hybrid Domains}\label{sec:diff_plan}

Differentiable planning formulates decision making as gradient-based optimization over a differentiable model of the dynamics. In deterministic settings \cite{wu2017scalable}, the transition model can be written as
\begin{equation}
    s_{t+1} = f(s_t, a_t),
    \label{eq:det_dynamics}
\end{equation}
and the objective in Eq.~\eqref{eq:mdp_objective} reduces to a deterministic function of the trajectory.

\paragraph{Deterministic differentiable planning.}
Planning can be posed as optimizing the action sequence $\mathbf{a} = (a_0, \dots, a_{H-1})$:
\begin{equation}
    \max_{\mathbf{a}} \; J(\mathbf{a})
    \quad \text{s.t.} \quad s_{t+1} = f(s_t, a_t).
\end{equation}
Unrolling Eq.~\eqref{eq:det_dynamics} yields a differentiable computation graph, enabling gradient-based optimization \cite{nocedal2006numerical} via backpropagation:
\begin{equation}
    \nabla_{a_t} J =
    \sum_{k=t}^{H-1}
    \frac{\partial r(s_k, a_k)}{\partial s_k}
    \frac{d s_k}{d a_t}
    + \frac{\partial r(s_t, a_t)}{\partial a_t},
\end{equation}
where sensitivities are computed recursively using the transition Jacobians:
\begin{equation}
    \frac{d s_{k+1}}{d s_k} = \frac{\partial f}{\partial s}(s_k, a_k),
    \quad
    \frac{d s_{k+1}}{d a_k} = \frac{\partial f}{\partial a}(s_k, a_k).
\end{equation}

\paragraph{Policy parameterization.}
Instead of optimizing actions directly, we optimize a parametric policy $a_t = \pi_\theta(s_t)$, where $\pi_\theta$ is a differentiable function (e.g., a neural network). This formulation, often referred to as deep reactive policies (DRP) \cite{bueno2019deep}, enables optimization over policy parameters via backpropagation through the system dynamics. This yields the objective
\begin{equation*}
    J(\theta) = \mathbb{E}\left[\sum_{t=0}^{H-1} r(s_t, \pi_\theta(s_t))\right].    
\end{equation*}
In deterministic settings, this reduces to a standard differentiable program where gradients $\nabla_\theta J$ are obtained by backpropagation through both the policy and the dynamics.

\paragraph{Stochastic dynamics and reparameterization.}
In stochastic domains, transitions depend on exogenous noise:
\begin{equation*}
    s_{t+1} \sim p(\cdot \mid s_t, a_t),
    \quad \mbox{or equivalently} \quad
    s_{t+1} = f(s_t, a_t, \xi_t), \;\; \xi_t \sim p(\xi).    
\end{equation*}
The objective becomes
\begin{equation*}
    J(\theta) = \mathbb{E}_{\xi}\left[\sum_{t=0}^{H-1} r(s_t, \pi_\theta(s_t))\right].    
\end{equation*}
Using reparameterization, the expectation is rewritten over noise variables independent of $\theta$, allowing gradients to pass through the trajectory:
\begin{equation*}
    \nabla_\theta J(\theta)
    = \mathbb{E}_{\xi}\left[
    \sum_{t=0}^{H-1}
    \frac{d r(s_t, \pi_\theta(s_t))}{d \theta}
    \right].
\end{equation*}
This transformation converts the stochastic computation graph into a fully differentiable one, enabling low-variance gradient estimation via standard backpropagation.\\
In practice, expectations are approximated using Monte Carlo rollouts:
\begin{equation*}
    \hat{J}(\theta) = \frac{1}{N} \sum_{i=1}^{N} \sum_{t=0}^{H-1} r(s_{i,t}, \pi_\theta(s_{i,t})),    
\end{equation*}
which defines the optimization objective used in stochastic gradient descent \cite{bueno2019deep}.

\paragraph{Hybrid discrete–continuous domains.}
In hybrid domains, the transition model includes discrete components and logical structure that are not directly differentiable. To enable gradient-based optimization, the system and reward are approximated by a differentiable surrogate:
\begin{equation*}
    s_{t+1} \approx \tilde{f}(s_t, a_t, \xi_t),    
\end{equation*}
where discrete operations are replaced by smooth relaxations, such as t-norms \cite{hajek2001metamathematics}.

For example, Boolean conjunction and conditional branching may be approximated as:
\begin{equation*}
    a \wedge b \approx T(a,b), \quad \mbox{if}\ c\ \mbox{then}\ a\ \mbox{else}\ b \approx c a + (1-c)b,    
\end{equation*}
yielding a fully differentiable computation graph over relaxed variables \cite{gimelfarb2024jaxplan}. 
However, this induces a fundamental trade-off: faithful approximations yield ill-conditioned objectives, while stronger relaxation makes optimization easier at the cost of distorting the underlying problem.

\begin{figure*}[t]
    \centering
    \begin{subfigure}{0.24\textwidth}
        \includegraphics[width=\linewidth]{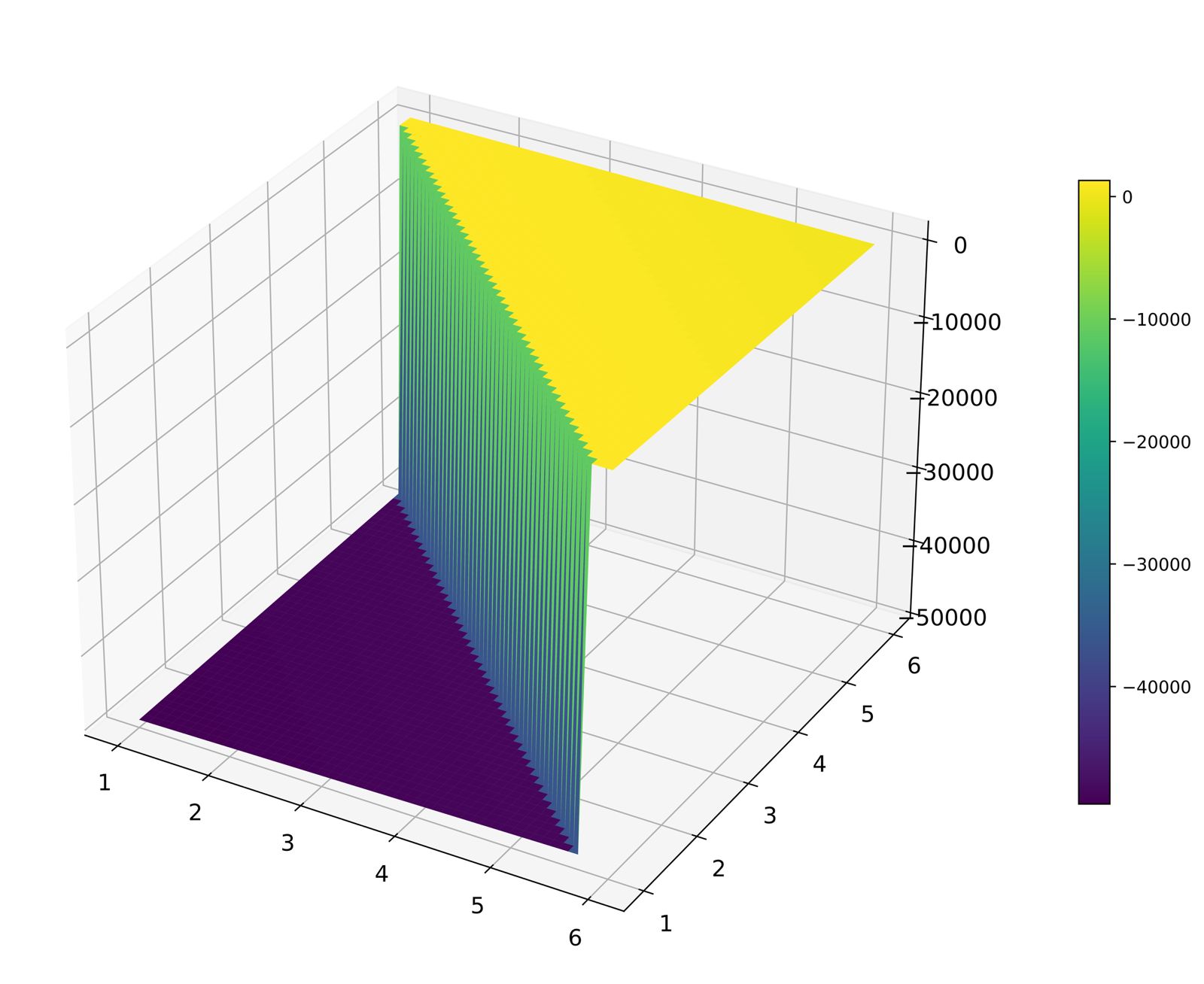}
        \caption{Exact: discontinuous landscape}
    \end{subfigure}
    \hfill
    \begin{subfigure}{0.24\textwidth}
        \includegraphics[width=\linewidth]{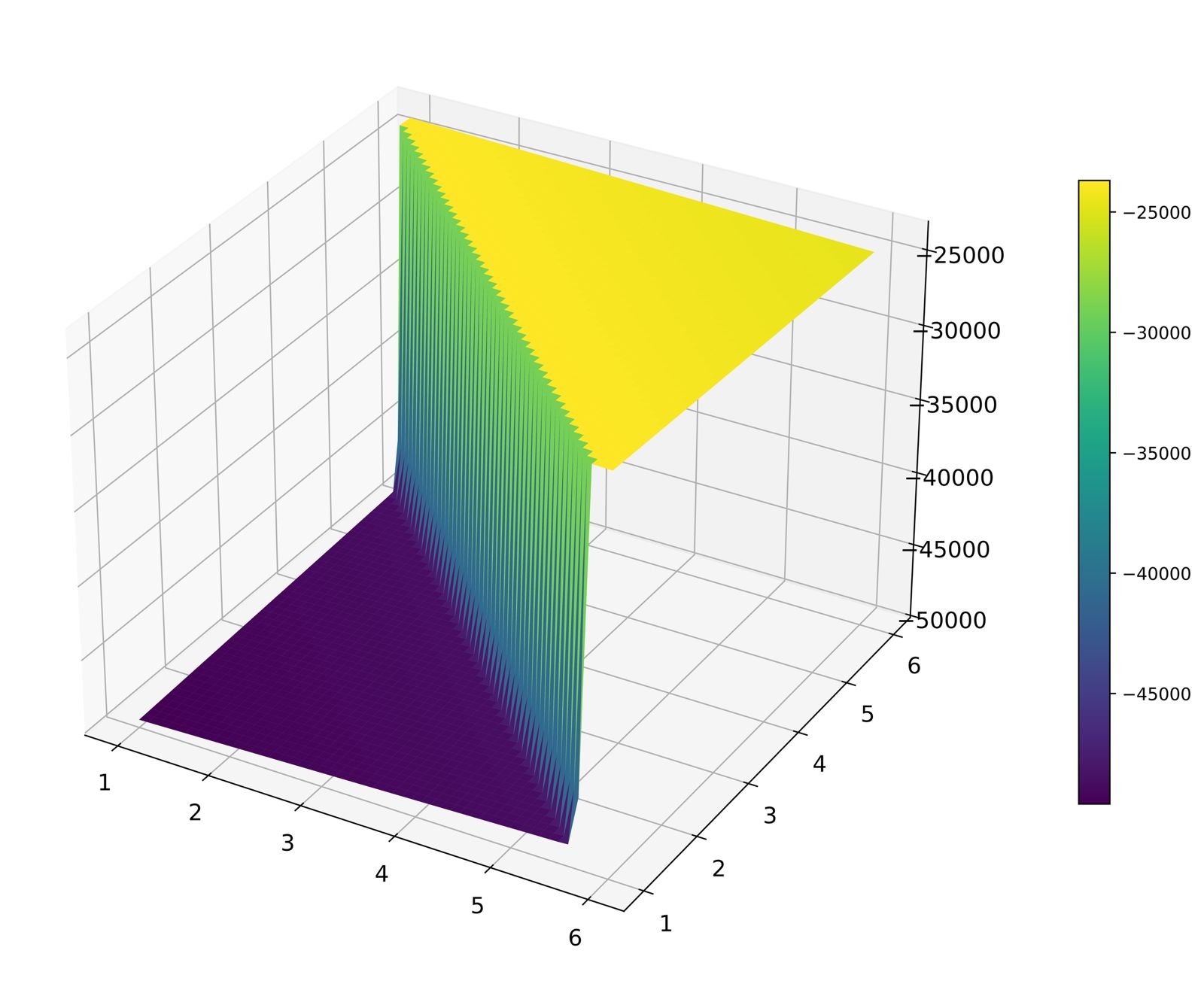}
        \caption{$w=50$: sharp but differentiable}
    \end{subfigure}
    \hfill
    \begin{subfigure}{0.24\textwidth}
        \includegraphics[width=\linewidth]{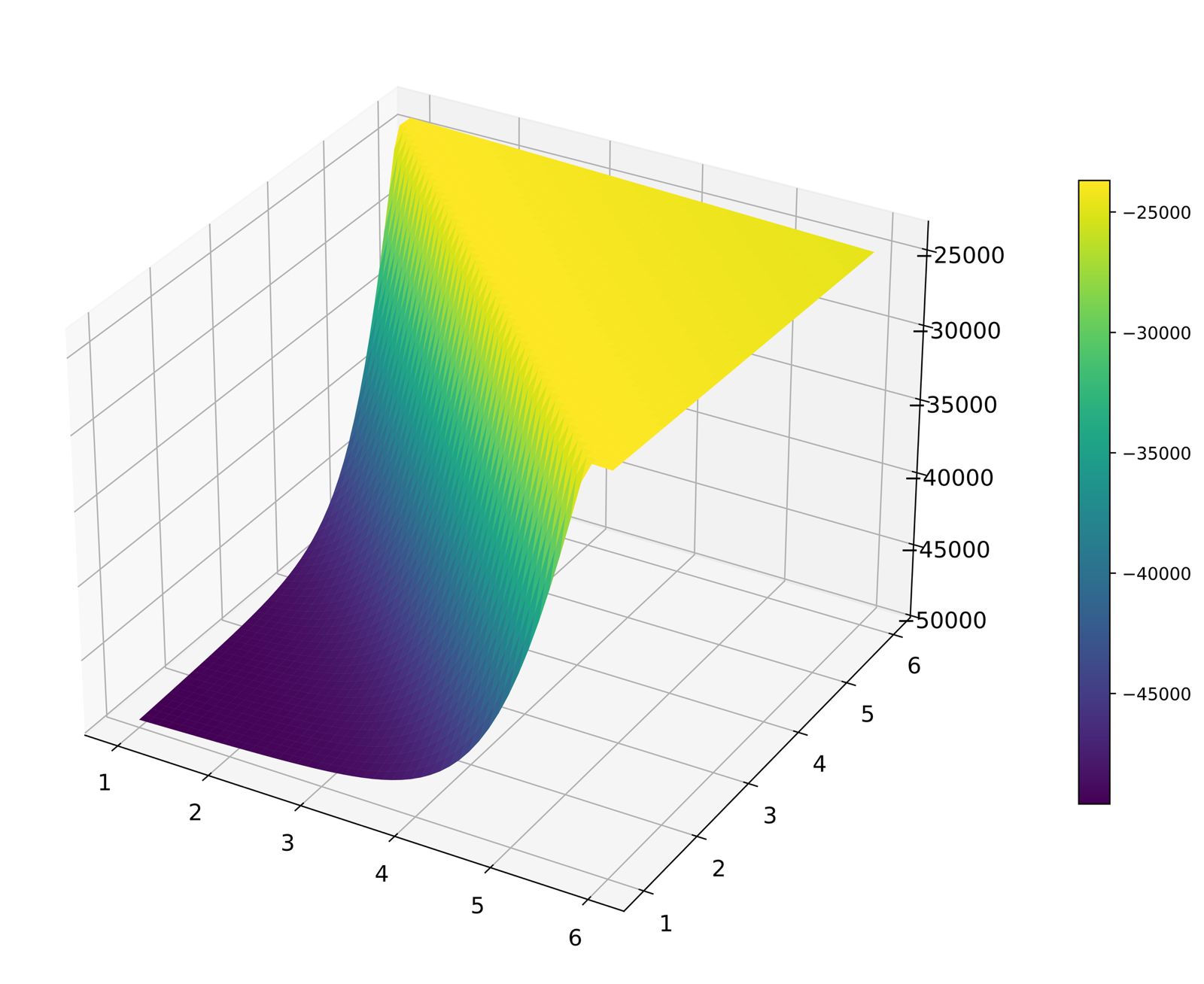}
        \caption{$w=2$: smoother, distorted}
    \end{subfigure}
    \hfill
    \begin{subfigure}{0.24\textwidth}
        \includegraphics[width=\linewidth]{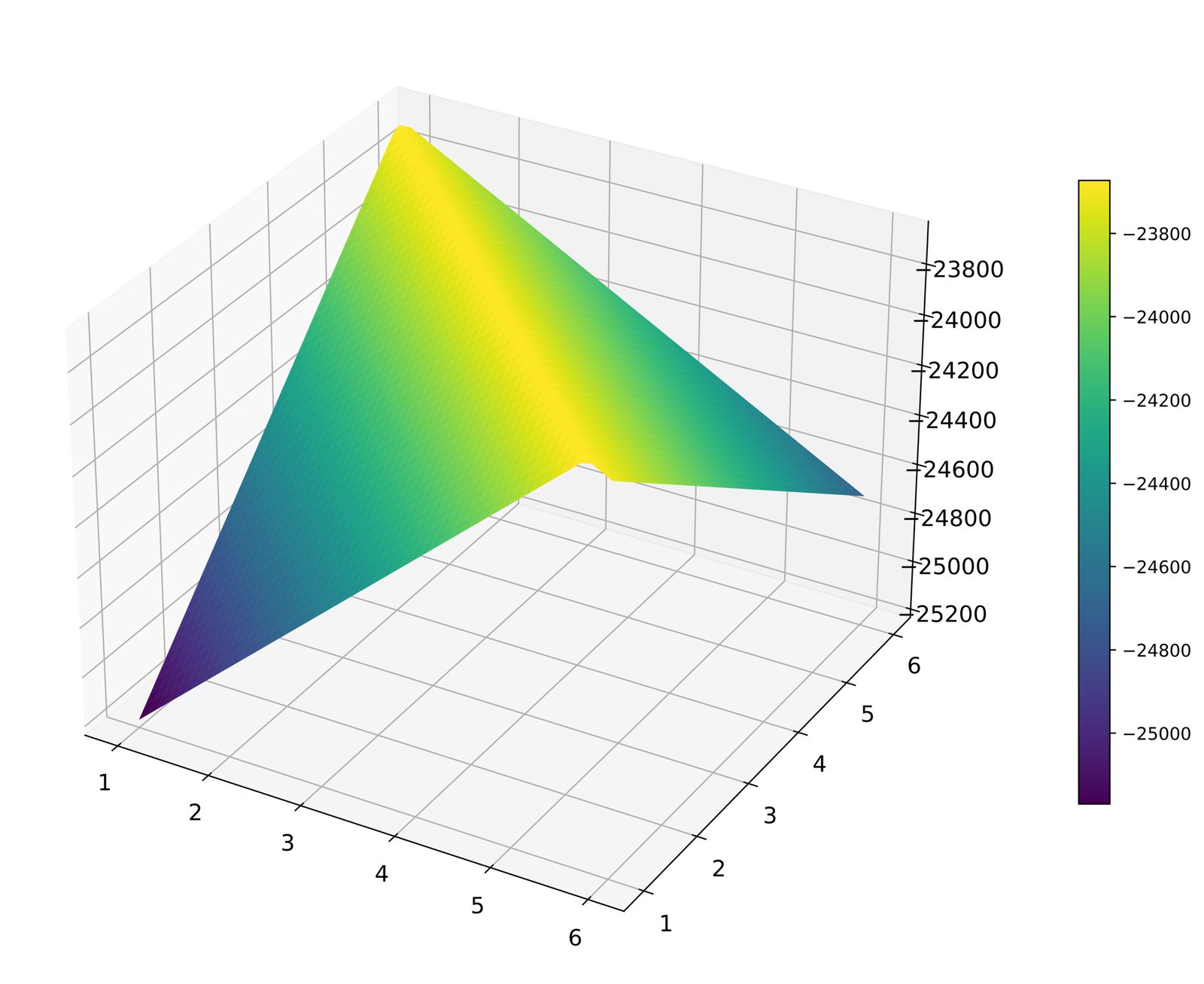}
        \caption{$w=0.01$: fully smooth, low fidelity}
    \end{subfigure}

    \caption{
    Optimization landscapes in the \texttt{PowerGen} domain under different levels of relaxation. 
    The exact dynamics (left) produce a non-smooth objective with discontinuities and flat regions. 
    Increasing relaxation smooths the surface but introduces a trade-off between optimization tractability and fidelity to the original problem.
    }
    \label{fig:powergen_landscapes}
\end{figure*}
\subsection{Optimization Landscape Pathologies}

We now analyze the optimization landscape induced by the formulation in Section~\ref{sec:diff_plan} and highlight key challenges that arise in hybrid domains. In particular, we focus on the effects of differentiable relaxations on the geometry of the objective function and their implications for gradient-based optimization.

\paragraph{Motivating example: Power generation domain.}
The \texttt{Power generation} (\texttt{PowerGen}) domain is a continuous relaxation of the classical unit commitment problem presented in the 2023 International Planning Competition (IPC) \cite{taitler2024ipc}, where the objective is to meet demand using a set of generators while minimizing operational costs. Each generator is controlled by a continuous action representing its production level, subject to capacity constraints and cost functions that include fixed and variable components. The underlying dynamics include threshold-based and logical conditions (e.g., activation costs and feasibility constraints), which introduce non-smooth behavior in the exact formulation. In this example, we consider two generators with actions $a^{(1)}, a^{(2)} \in [1,6]$, and evaluate constant policies over a fixed horizon, allowing us to visualize the induced objective landscape of $J(a^{(1)}, a^{(2)})$ as a function of the constant action pair.

% \paragraph{From Exact dynamics to smooth distorted differentiable relaxations}
Figure~\ref{fig:powergen_landscapes}(a) shows the true objective surface under the exact (non-relaxed) dynamics. The landscape exhibits sharp discontinuities induced by underlying discrete decisions (e.g., activation thresholds and logical constraints), resulting in large flat regions separated by abrupt transitions. In such regions, gradients are either undefined or uninformative, rendering gradient-based optimization ineffective.
To enable gradient-based optimization, hybrid dynamics are approximated using smooth relaxations (Section~\ref{sec:diff_plan}), controlled by a parameter $w$ (e.g., sigmoid or t-norm sharpness). Figures~\ref{fig:powergen_landscapes}(b)--(d) illustrate the effect of varying this parameter.
For large values of $w$ (Figure~\ref{fig:powergen_landscapes}(b)), the relaxation closely approximates the original dynamics, preserving the overall structure of the objective. However, the resulting surface remains highly ill-conditioned, with steep transitions and extended flat regions, leading to poor gradient signal.
For moderate values (Figure~\ref{fig:powergen_landscapes}(c)), the surface becomes smoother and gradients are better behaved. However, this smoothing begins to alter the geometry of the objective, shifting optima and distorting the relative quality of solutions.
For small values (Figure~\ref{fig:powergen_landscapes}(d)), the landscape becomes fully smooth and easy to optimize, but significantly deviates from the true objective, potentially leading to solutions that are suboptimal or infeasible under the original dynamics.

\paragraph{Trade-off between fidelity and optimization.}
These observations highlight a fundamental trade-off: relaxations that closely approximate the original problem yield differentiable but poorly conditioned optimization landscapes, while more aggressive smoothing improves gradient behavior at the cost of fidelity to the true objective.
As a result, gradient-based optimization in differentiable planning is inherently challenging. Optimizers may become trapped in flat regions, fail to traverse sharp transitions, or converge to solutions that are optimal for the relaxed problem but suboptimal for the true dynamics. These challenges persist even when the underlying model is accurate and the relaxation is carefully tuned.
This motivates the need for optimization strategies that explicitly account for the structure of the landscape and enable more effective exploration of the action space.

\section{Model-Driven Policy Optimization via Stochastic Exploration}

We introduce \textit{Model-Driven Policy Optimization} (MDPO), a framework that augments differentiable planning with stochastic exploration in the action space. 
We consider a parametric policy $\pi_\theta$ (implemented as a neural network), and optimize its parameters via gradient-based methods.
The key idea is to replace deterministic policy execution with a stochastic perturbation, enabling improved exploration of the optimization landscape while preserving end-to-end differentiability. Unlike standard stochastic optimization techniques that inject isotropic noise, MDPO introduces structured stochasticity guided by the planning dynamics and model structure. We refer to this as model-driven stochasticity because perturbations are applied at the level of action trajectories within the planning model, ensuring that exploration respects system dynamics rather than operating directly in parameter space. Specifically, perturbations are applied in a way that preserves trajectory feasibility while encouraging exploration across discontinuities in the optimization landscape.

\subsection{Stochastic Action Perturbation}
Given a parametric policy $\pi_\theta(s_t)$, standard differentiable planning executes actions deterministically:
\begin{equation*}
    a_t = \pi_\theta(s_t).    
\end{equation*}

In MDPO, we instead define a stochastic policy by injecting additive noise:
\begin{equation*}
    \tilde{a}_t = \tilde{\pi}_\theta(s_t) = \pi_\theta(s_t) + \epsilon_t, 
    \quad \epsilon_t \sim \mathcal{N}(0, \Sigma_t),
\end{equation*}
where $\Sigma$ is the covariance matrix controlling the magnitude and structure of the perturbation.

The system then evolves according to the (possibly stochastic or relaxed) dynamics:
\begin{equation*}
    s_{t+1} = f(s_t, \tilde{a}_t, \xi_t).    
\end{equation*}

\paragraph{Objective under stochastic policy.}
Under this formulation, the optimization objective becomes the expected cumulative reward under the perturbed policy:
\begin{equation*}
    J(\theta) = \mathbb{E}_{\epsilon, \xi} \left[ \sum_{t=0}^{H-1} r(s_t, \tilde{a}_t) \right]
    % J(\theta) = \mathbb{E}_{\tilde a_t\sim \tilde \pi} \left[ \sum_{t=0}^{H-1} r(s_t, \tilde{a}_t) \right].    
\end{equation*}
In practice, this expectation is approximated using Monte Carlo rollouts.

%%%%%%%%%%%

\paragraph{Gradient estimation.}
Using the reparameterization $\tilde{a}_t = \pi_\theta(s_t) + \epsilon_t$, where $\epsilon_t$ is independent of $\theta$, gradients can be computed by backpropagation through the stochastic computation graph:
\begin{equation*}
\nabla_\theta J(\theta)
= \mathbb{E}_{\epsilon, \xi} \left[
\nabla_\theta \sum_{t=0}^{H-1} r(s_t, \tilde{a}_t)
\right].
\end{equation*}
Since $\epsilon_t$ does not depend on $\theta$, it does not contribute to the gradient, and standard automatic differentiation can be applied.

\begin{thm} \label{thm:grad}
Under the reparameterized dynamics, the gradient $\nabla_\theta J(\theta)$ can be computed by backpropagation through the stochastic computation graph, and is equivalent to differentiating the cumulative reward along sampled trajectories.
\begin{equation*}
    \nabla_\theta J(\theta)
    =
    \mathbb{E}_{\epsilon,\xi} \left[
    \sum_{t=0}^{H-1}
    \frac{\partial r}{\partial \tilde{a}_t}
    \frac{\partial \pi_\theta}{\partial \theta}
    +
    \sum_{t=0}^{H-1}
    \frac{\partial r}{\partial s_t}
    \frac{\partial s_t}{\partial \theta}
    \right]
\end{equation*}
\end{thm}
A full expansion of the gradient in terms of the system Jacobians is provided in the appendix.
%%%%%%%%%

The stochastic policy $\tilde{\pi}_\theta$ induces exploration directly in the action space, allowing the optimizer to probe nearby regions of the objective landscape. Such exploration is particularly important in hybrid domains, where flat regions and sharp transitions can hinder deterministic gradient-based optimization. This stochastic perturbation enables the optimizer to explore directions that would be inaccessible under deterministic gradients, particularly in flat or saturated regions of the objective landscape. As a result, the method can escape poor local optima and discover higher-quality solutions. In this context, the covariance $\Sigma$ becomes a domain-dependent parameter that governs the extent and structure of exploration, and must be chosen to effectively navigate the underlying optimization landscape.

The introduction of stochastic perturbations induces a trade-off. For small values of $\Sigma$, the perturbed policy remains close to the original deterministic policy, and the resulting objective closely approximates the original optimization problem. As $\Sigma$ increases, the sampling distribution deviates further from the deterministic policy, introducing bias with respect to the original objective. At the same time, the gradient estimator becomes higher variance due to increased stochasticity in the sampled trajectories. Thus, the choice of $\Sigma$ plays a critical role in balancing exploration and optimization stability.

\paragraph{Extension to discrete actions.}
While the formulation above assumes continuous action spaces, the same principle can be extended to discrete settings by introducing stochastic relaxations or sampling mechanisms (e.g., via softmax or Gumbel-based perturbations), enabling analogous exploration in hybrid or discrete domains.

\subsection{Adaptive Noise Scaling}

Since the planning model is fully available, we can leverage it to derive informative exploration signals directly from the optimization landscape. We propose an adaptive mechanism for controlling the magnitude of action-space exploration based on the local sensitivity of the trajectory objective. This results in a time-dependent noise profile $\{\sigma_t\}_{t=0}^{H-1}$.

We propose a two-pass procedure in which each iteration consists of two rollouts. First, an \emph{analysis rollout} is generated using the current policy without action noise. This rollout is used to estimate the sensitivity of the objective with respect to actions along the trajectory. Based on this analysis, a timestep-dependent noise profile is constructed. Then, a second \emph{update rollout} is performed using the derived noise profile to update the policy parameters. This separates the estimation of the optimization landscape from the stochastic optimization step.

We consider stochastic dynamics of the form
\begin{equation*}
    s_{t+1} = f(s_t, a_t, \xi_t), \quad \xi_t \sim p(\xi),
\end{equation*}
where $\xi_t$ represents exogenous randomness that is not controlled by the policy. A trajectory induced by a deterministic policy and a realization of $\xi_{0:H-1}$ is
\begin{equation*}
    \tau(\theta, \xi) = (s_0, a_0, s_1, a_1, \dots, s_H, \xi_0,\dots, \xi_{H-1}), \quad a_t = \pi_\theta(s_t) \in \mathbb{R}^n.
\end{equation*}
The objective is the expected cumulative reward: 
\begin{equation*}
  J(\theta) = \mathbb{E}_{\xi}\left[\sum_{t=0}^{H-1} r(s_t, \pi_\theta(s_t))\right].  
\end{equation*}
Using the reparameterized dynamics, gradients can be propagated through the trajectory while treating $\xi_t$ as fixed samples. Given an analysis trajectory $\tau(\theta, \xi)$, we compute, for each timestep $t$, the multi-dimensional gradient of the vector $a_t\in\mathbb{R}^n$:
\begin{equation*}
% g_t = \nabla_{a_t} \sum_{k=0}^{H-1} r(s_k, a_k) 
% \quad \rightarrow \quad g_t = \Big( \frac{\partial J}{\partial a_t^{(1)}},..., \frac{\partial J}{\partial a_t^{(n)}}  \Big)
    g_t = \nabla_{a_t} \sum_{k=0}^{H-1} r(s_k, a_k)
        = \left( \frac{\partial J}{\partial a_t^{(1)}}, \dots, \frac{\partial J}{\partial a_t^{(n)}} \right)
\end{equation*}
which corresponds to a Monte Carlo estimate of $\nabla_{a_t} J(\theta)$. We define a scalar sensitivity score:
\begin{equation*}
    s_t = \| g_t \|_2.
\end{equation*}
In principle, the expectation over $\xi$ can be estimated using multiple rollouts. In practice, we use a single trajectory sample per iteration, which was found sufficient in our experiments.
To obtain comparable values across the trajectory, sensitivities are normalized using a high quantile:
\begin{equation*}
    \hat{s}_t = \mathrm{clip}\!\left(
    \frac{s_t}{q_p + \lambda}, 0, 1
    \right),
    \quad
    q_p = \mathrm{Quantile}_p(\{s_t\}),
\end{equation*}
where $\lambda$ is a small regularization term to avoid division by zero, and $p=0.95$ in this work.
The normalized sensitivity is mapped to a timestep-dependent noise scale:
\begin{equation*}
    \sigma_t = \sigma_{\max} + (\sigma_{\min} - \sigma_{\max})(1 - \hat{s}_t)^\alpha,
\end{equation*}
where $\sigma_{\min}$, $\sigma_{\max}$, and $\alpha$ control the range and shape of the mapping.
The resulting profile is used in the update rollout:
\begin{equation*}
\tilde{a}_t = \pi_\theta(s_t) + \eta_t,  \quad \eta_t \sim \mathcal{N}(0, \sigma_t^2 I),
\end{equation*}
and the policy parameters are updated using the resulting trajectory.
This mapping assigns larger noise to timesteps with higher sensitivity, encouraging exploration where actions have greater impact on the trajectory objective.

\section{Experiments}
\paragraph{Benchmark Problems}
We evaluate the proposed method on three benchmark domains from the probabilistic planning and RL track of 2023 IPC \cite{taitler2024ipc}: \texttt{PowerGen} \cite{padhy2004unit}, \texttt{Reservoir Control} \cite{yeh1985reservoir}, and \texttt{HVAC} \cite{tiugrek2002nonlinear}. These domains exhibit nonlinear and hybrid discrete–continuous dynamics, making them suitable testbeds for differentiable planning. For each domain, we consider two instances, resulting in six evaluation settings. A detailed description is provided in the appendix. \\
\noindent\hspace*{0.75em}\texttt{PowerGen.}
A continuous relaxation of the unit commitment problem, where generators must meet stochastic demand while minimizing production and switching costs. The domain combines continuous control with threshold-based activation and nonlinear demand. We consider instances with 7 and 11 generators. \\
\noindent\hspace*{0.75em}\texttt{HVAC.}
A nonlinear control problem over interconnected zones, where actions regulate airflow and heating to maintain comfort while minimizing energy cost. The dynamics involve nonlinear heat transfer and coupling between zones. We consider instances with 10 and 20 zones. \\
\noindent\hspace*{0.75em}\texttt{Reservoir Control.}
A network of reservoirs with stochastic inflow, evaporation, and flow constraints. Actions control water release, with penalties for deviations from desired levels. We consider instances with 15 and 30 reservoirs.

\textbf{Methods and Baselines.}
We compare MDPO against several baselines to isolate the effect of stochastic exploration. \\
\noindent\hspace*{0.75em}\textit{Deterministic differentiable planning.}
A noise-free version of our method ($\Sigma = 0$), sharing the same model, policy parameterization, and optimizer. This baseline isolates the effect of stochastic exploration. \\
\noindent\hspace*{0.75em}\textit{JaxPlan.}
A publicly available implementation of differentiable planning based on relaxed hybrid dynamics. This serves as a strong optimization baseline representative of prior work \cite{gimelfarb2024jaxplan}. \\
\noindent\hspace*{0.75em}\textit{MDPO (Constant noise).}
We consider fixed Gaussian noise with $\sigma \in \{1,3\}$ applied to the action space during optimization. These variants evaluate the effect of non-adaptive stochastic exploration. These values were chosen as representative moderate noise levels that provide consistent improvements while avoiding instability, as larger noise magnitudes can lead to distributional drift and degrade optimization.\\
\noindent\hspace*{0.75em}\textit{MDPO (Adaptive).}
Our method uses a time-dependent noise profile derived from gradient-based sensitivity. In all experiments, we set $\sigma_{\min}=0$ and a large $\sigma_{\max}$ (typically in the range $[9,11]$), providing a wide exploration range and allowing the method to flexibly adapt noise levels to the local optimization landscape. Results were not highly sensitive to the exact choice of $\sigma_{\max}$, provided it remained sufficiently large.\\
\noindent\hspace*{0.75em}\textit{Model-free RL.}
We include Proximal Policy Optimization (PPO) \cite{schulman2017proximal} as a model-free baseline. PPO uses the same policy architecture, horizon, and initialization, and performs 10 updates per rollout.

\textbf{Implementation Details.}
All differentiable planning methods operate on models using t-norm relaxations with weight $w=100$, providing a close approximation to the original dynamics. MDPO was implemented in PyTorch. PPO, as a model-free baseline, is trained directly on the true (non-relaxed) environment.
Policies are parameterized by feedforward neural networks with two hidden layers of 12 neurons each, ensuring that performance differences reflect optimization rather than representational capacity. All methods are trained for 3000 iterations using RMSProp \cite{hinton2012rmsprop} (learning rate $0.01$) with horizon $H=120$, where each iteration consists of a single rollout and gradient update. All methods share the same initialization, training budget, and optimization settings.
Experiments are conducted over 20 seeds, with evaluations averaged over 20 rollouts on the true (non-relaxed) dynamics in an external pyRDDLGym \cite{taitler2022pyrddlgym} implementation. All experiments were conducted on a compute cluster, with each run allocated 2 CPUs and 8GB of memory.

\textbf{Overall Performance.}
Table~\ref{tab:results} summarizes final performance across all benchmark instances. PPO performs poorly in all domains, showing little improvement within the given budget, highlighting the difficulty of learning without a model. Deterministic differentiable planning and JaxPlan achieve similar results, consistently converging to suboptimal solutions.
Introducing stochasticity via fixed action noise improves performance, but remains sensitive to the choice of scale and varies across domains. In contrast, the adaptive noise method consistently achieves the best performance across all instances.
In addition, MDPO exhibits low variance across seeds while maintaining high solution quality. PPO shows low variance only due to its lack of improvement, resulting in nearly constant returns. In \texttt{HVAC20}, although MDPO does not have the lowest variance, it is the only method that achieves meaningful optimization, while others remain significantly worse.
These results empirically support the hypothesis that stochastic exploration mitigates optimization pathologies arising from flat and ill-conditioned regions of the objective landscape.

\begin{table}[t]
\centering
\scriptsize
\resizebox{\textwidth}{!}{
\begin{tabular}{l cc cc cc}
\toprule
 & \multicolumn{2}{c}{\texttt{HVAC}} 
 & \multicolumn{2}{c}{\texttt{PowerGen}}
 & \multicolumn{2}{c}{\texttt{Reservoir}} \\
\cmidrule(lr){2-3} \cmidrule(lr){4-5} \cmidrule(lr){6-7}
Method & 10 & 20 & 7 & 11 & 15 & 30 \\
\midrule

PPO 
& -4,706,559.94 $\pm$ 0.38 & -5,799,263.08 $\pm$ 0.21
& -119,950.01 $\pm$ 217.92 & -120,000.00 $\pm$ 0.00 
& -1,567,404.88 $\pm$ 258.54 & -3,821,075.16 $\pm$ 226.01 \\

Deterministic  
& -3,268,060.37 $\pm$ 441,267.23 & -5,799,274.81 $\pm$ 44.65 
& -38,565.41 $\pm$ 25,582.69 & -57,528.37 $\pm$ 30,444.41 
& -683,721.06 $\pm$ 221,923.57 & -2,219,640.67 $\pm$ 296,596.36 \\

JaxPlan 
& -2,987,355.11 $\pm$ 979,087.04 & -5,793,011.93 $\pm$ 89,429.51 
& -44,747.36 $\pm$ 33,219.89 & -54,034.04 $\pm$ 24,745.46 
& -688,683.95 $\pm$ 205,670.07 & -2,364,420.45 $\pm$ 388,492.15 \\

\addlinespace[1pt]
\cdashline{1-7}
\addlinespace[2pt]

Constant 1  
& -1,693,569.37 $\pm$ 129,419.57 & -5,799,302.46 $\pm$ 17.11
& 6,753.42 $\pm$ 5,279.33 & 16,345.46 $\pm$ 3,977.71 
& -249,675.61 $\pm$ 20,927.82 & -1,217,795.17 $\pm$ 151,823.21 \\

Constant 3 
& -1,555,612.12 $\pm$ 47,196.07 & -5,799,423.99 $\pm$ 58.96 
& 4,622.72 $\pm$ 7,027.73 & 13,772.41 $\pm$ 5,851.97
& -248,132.79 $\pm$ 21,170.60 & -1,115,865.90 $\pm$ 69,979.64 \\

Adaptive 
& \textbf{-1,371,468 $\pm$ 194.35} & \textbf{-2,086,875.93 $\pm$ 42,832.24}
& \textbf{10,076 $\pm$ 3,925.23} & \textbf{16,890.15 $\pm$ 3,138.48 }
& \textbf{-246,613.49 $\pm$ 18,382.48} & \textbf{-815,510.27 $\pm$ 59,029.69} \\

\bottomrule
\vspace{0.01mm}
\end{tabular}
}
\caption{
\textbf{Final performance across benchmark domains and instances.}
Results are averaged over 20 seeds and 20 evaluation rollouts per seed on the true dynamics, using the mean performance over the last 5 optimization iterations.
Adaptive noise consistently outperforms deterministic planning, JaxPlan, and constant-noise variants, while PPO performs poorly.
}
\label{tab:results}
\end{table}

\begin{figure}[t]
    \centering
        \resizebox{\linewidth}{!}{%
\footnotesize
\tikz[baseline=-0.6ex]{\draw[green!60!black, line width=1pt] (0,0)--(0.35,0);} Adaptive \hspace{0.8em}
\tikz[baseline=-0.6ex]{\draw[blue, dashed, line width=1pt] (0,0)--(0.35,0);} Deterministic ($\sigma=0$) \hspace{0.8em}
\tikz[baseline=-0.6ex]{\draw[black, dashed, line width=1pt] (0,0)--(0.35,0);} Const.\ noise ($\sigma=1$) \hspace{0.8em}
\tikz[baseline=-0.6ex]{\draw[red, dashed, line width=1pt] (0,0)--(0.35,0);} Const.\ noise ($\sigma=3$) \hspace{0.8em}
\tikz[baseline=-0.6ex]{\draw[black, dotted, line width=1pt] (0,0)--(0.35,0);} JaxPlan \hspace{0.8em}
\tikz[baseline=-0.6ex]{\draw[orange, dotted, line width=1pt] (0,0)--(0.35,0);} PPO
}
    \begin{subfigure}{0.32\textwidth}
        \includegraphics[width=\linewidth]{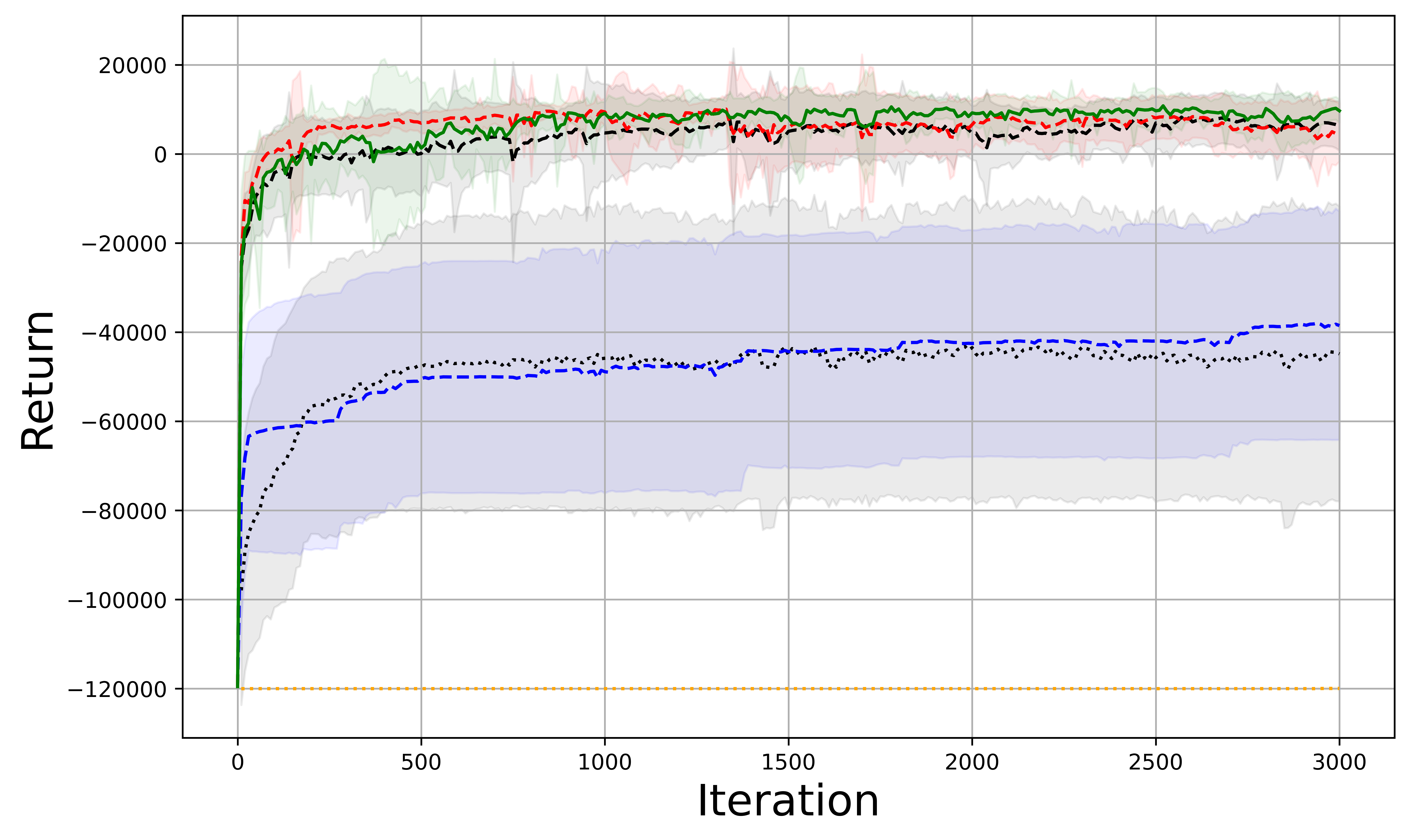}
        \caption{\texttt{PowerGen7}}
        \label{fig:learning_curves:p4}
    \end{subfigure}
    \hfill
    \begin{subfigure}{0.32\textwidth}
        \includegraphics[width=\linewidth]{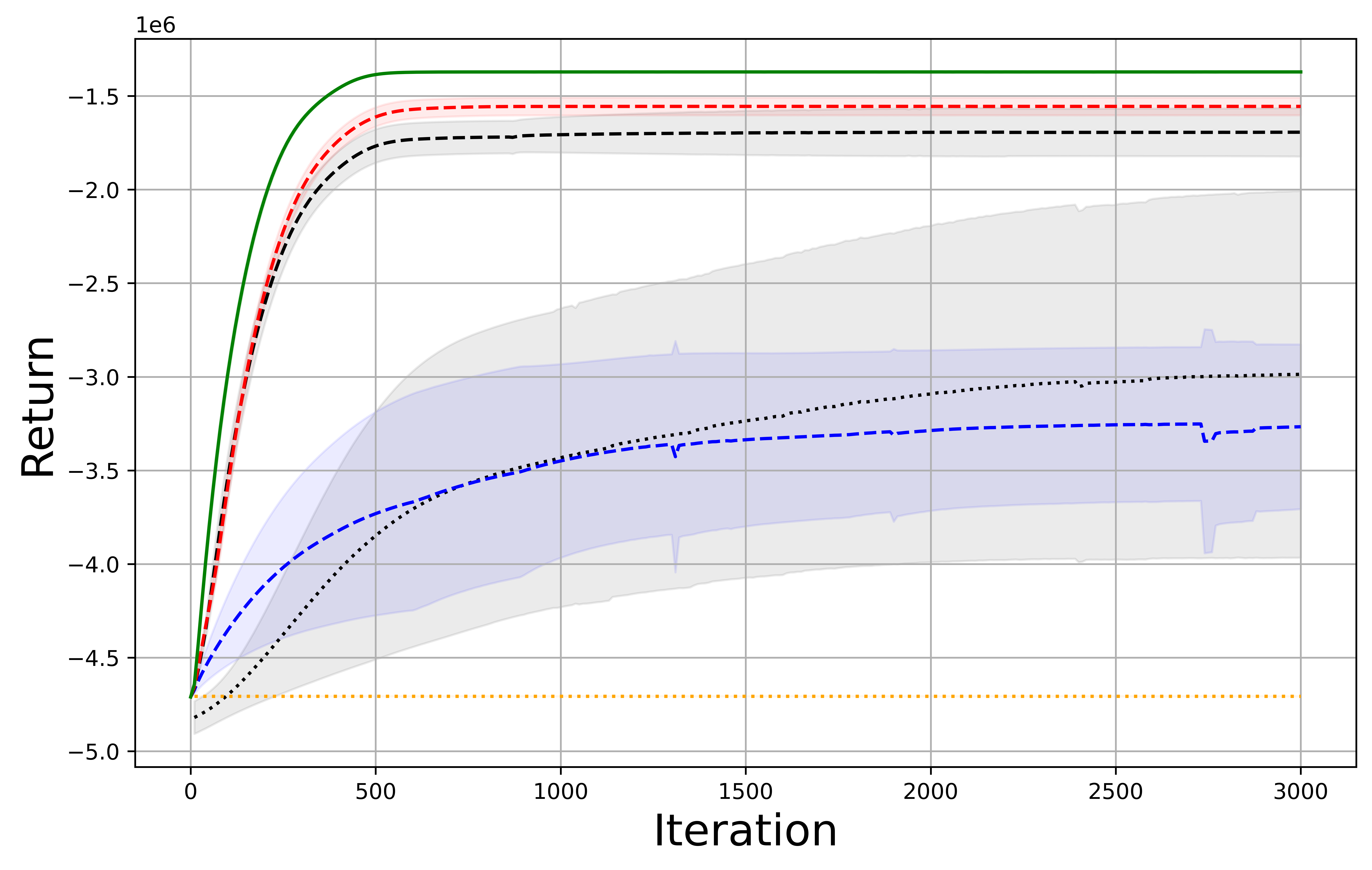}
        \caption{\texttt{HVAC10}}
        \label{fig:learning_curves:h4}
    \end{subfigure}
    \hfill
    \begin{subfigure}{0.32\textwidth}
        \includegraphics[width=\linewidth]{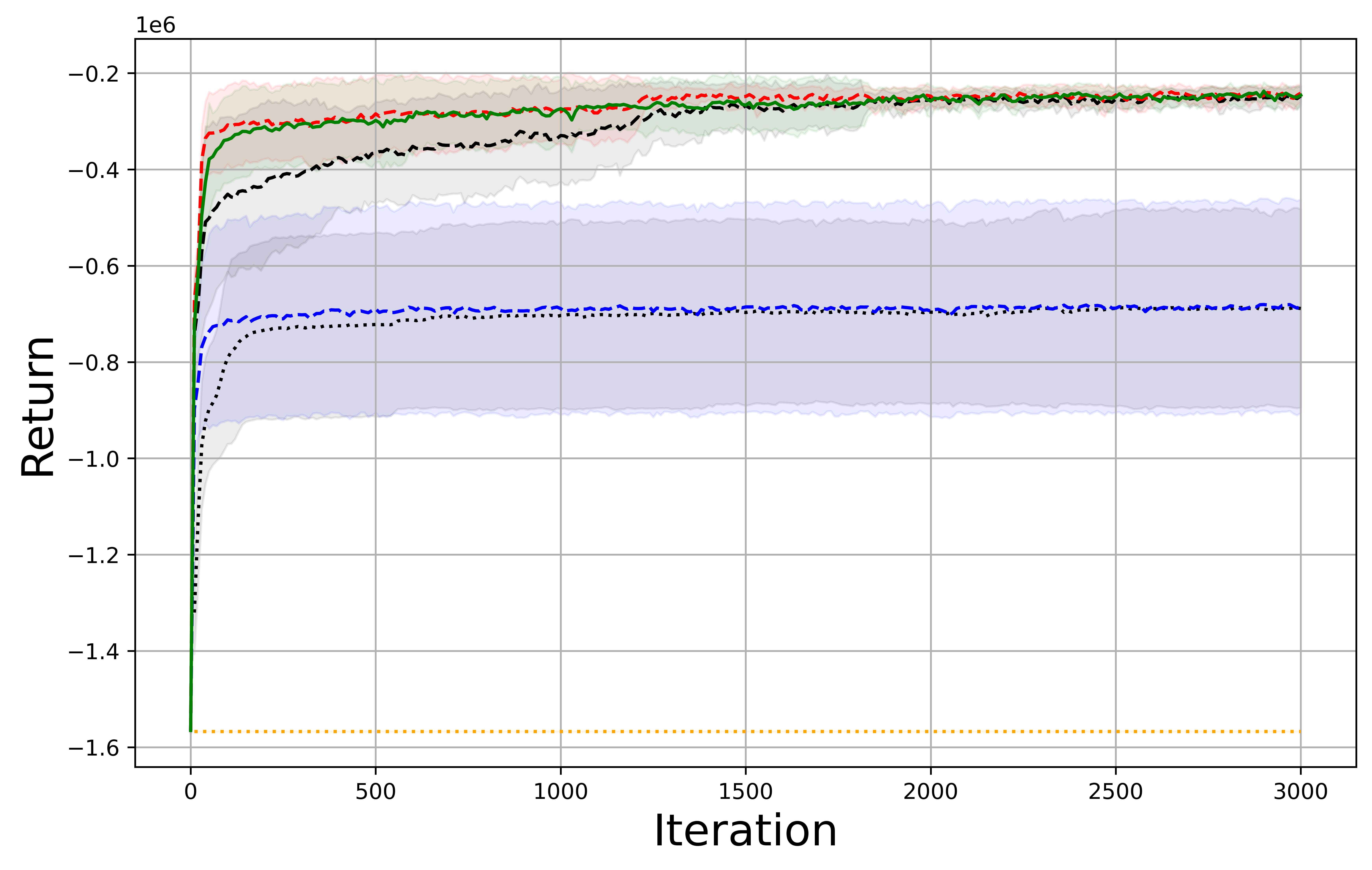}
        \caption{\texttt{Reservoir15}}
        \label{fig:learning_curves:r4}
    \end{subfigure}
    \hfill
    \begin{subfigure}{0.32\textwidth}
        \includegraphics[width=\linewidth]{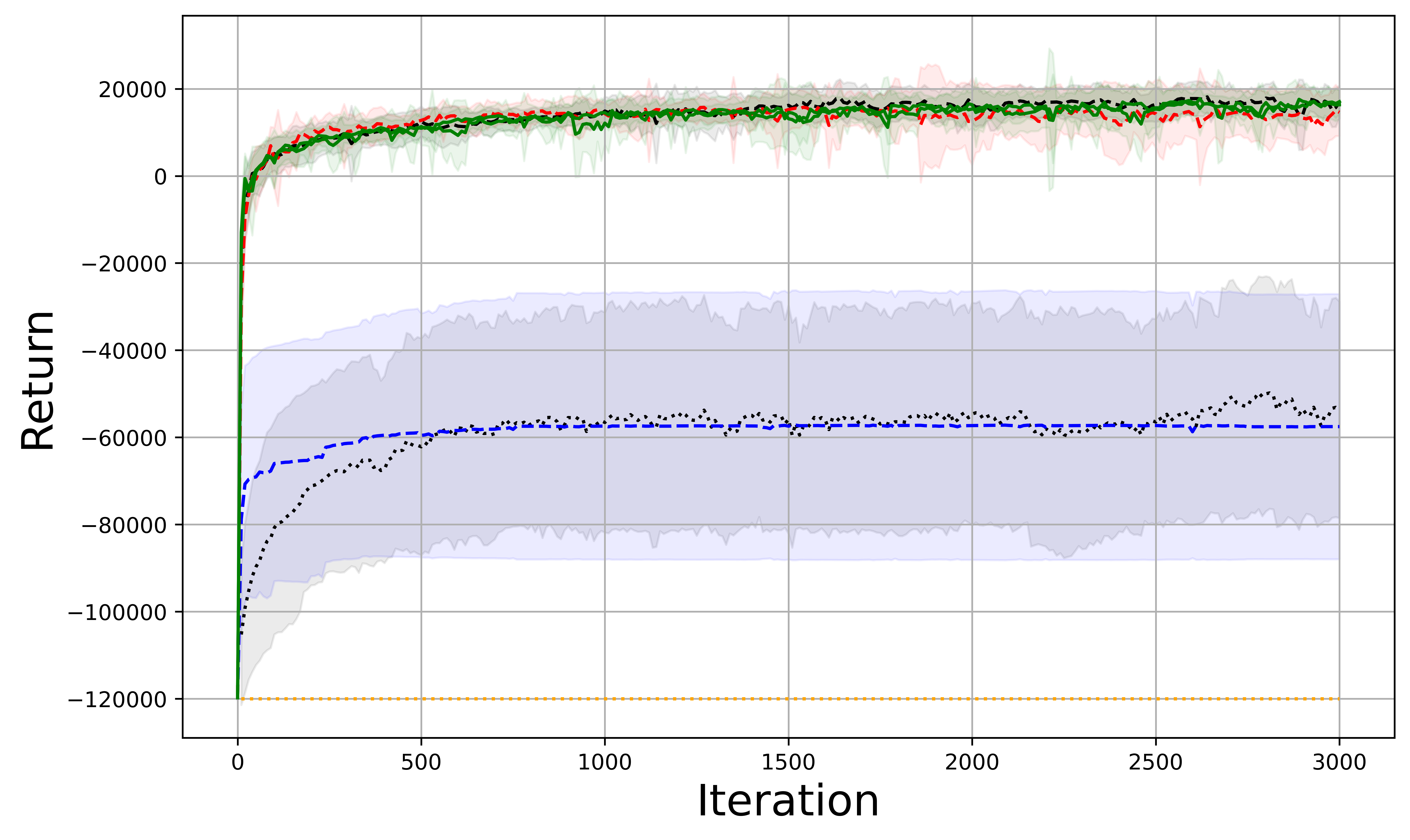}
        \caption{\texttt{PowerGen11}}
        \label{fig:learning_curves:pg5}
    \end{subfigure}
    \hfill
    \begin{subfigure}{0.32\textwidth}
        \includegraphics[width=\linewidth]{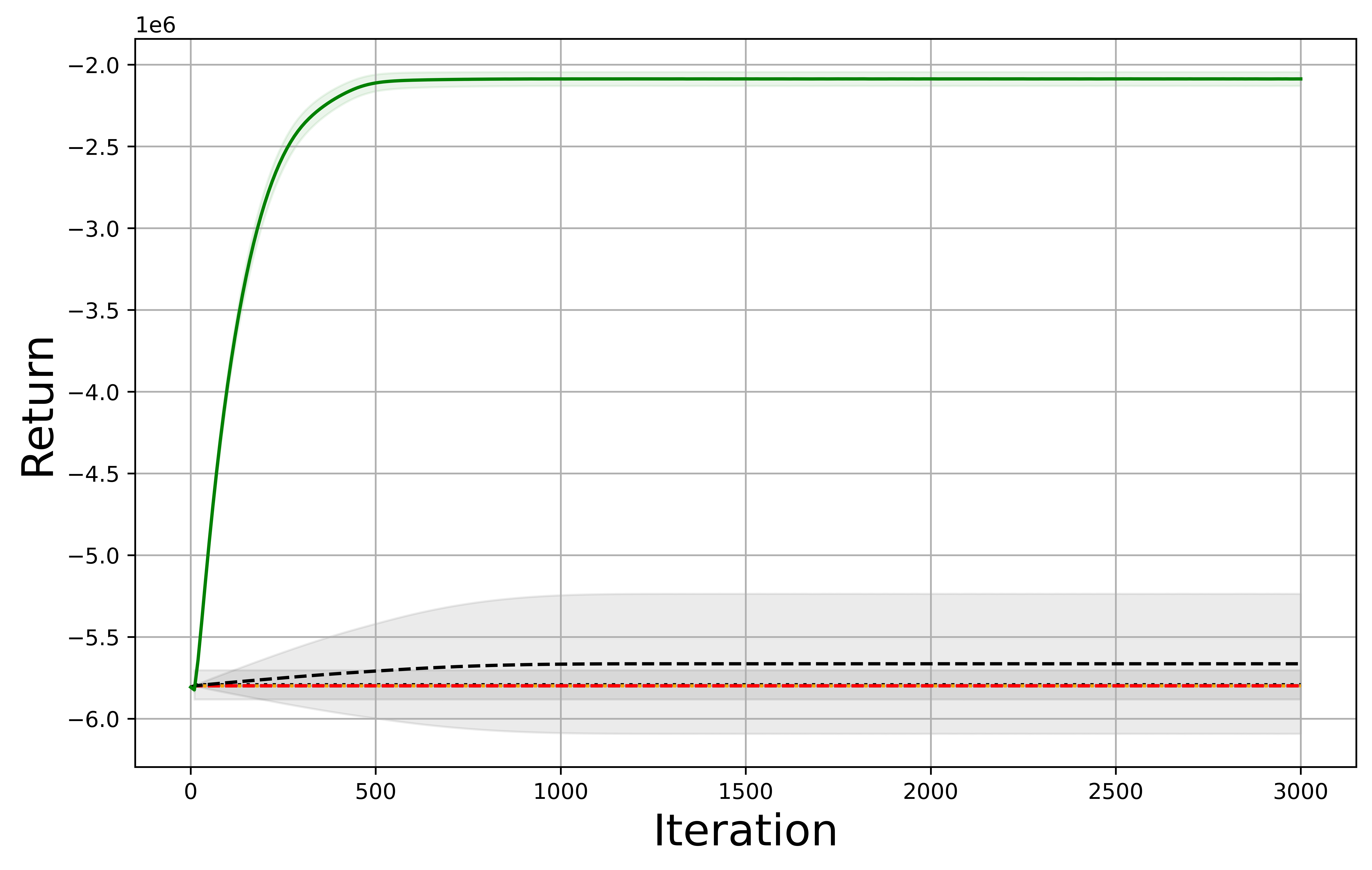}
        \caption{\texttt{HVAC20}}
        \label{fig:learning_curves:h5}
    \end{subfigure}
    \hfill
    \begin{subfigure}{0.32\textwidth}
        \includegraphics[width=\linewidth]{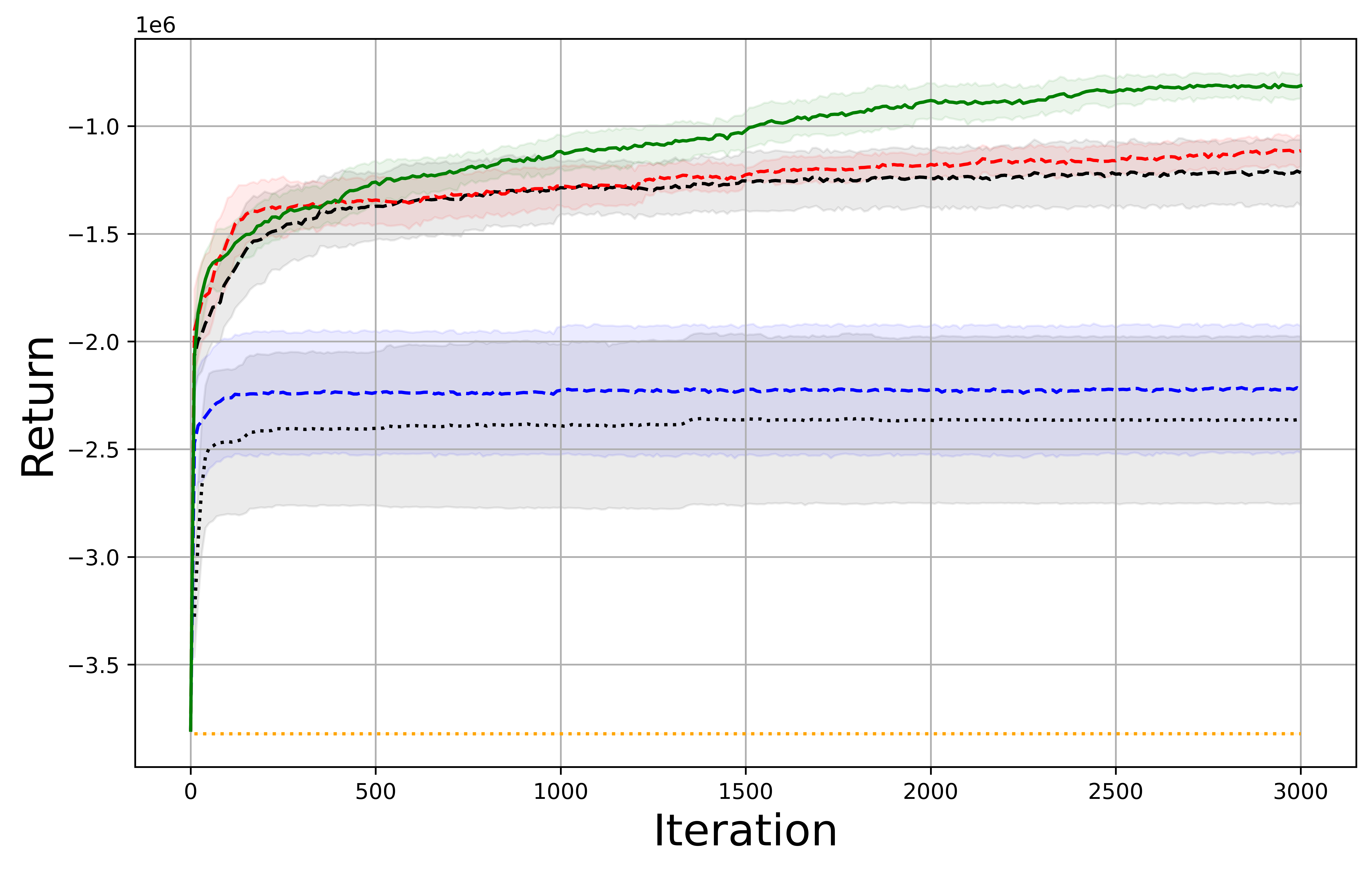}
        \caption{\texttt{Reservoir30}}
        \label{fig:learning_curves:r5}
    \end{subfigure}
    \caption{
        \textbf{Learning curves across benchmark instances.}
        Performance over 3000 optimization iterations for all methods, averaged over 20 random seeds (shaded regions indicate standard deviation).
        Deterministic differentiable planning and JaxPlan quickly plateau, while PPO shows limited improvement under the given optimization budget.
        Introducing stochastic exploration improves performance, with constant noise yielding moderate gains.
        The proposed adaptive method consistently achieves faster convergence and higher final performance, with the advantage becoming more pronounced in harder instances.
    }
    \label{fig:learning_curves}
\end{figure}

\paragraph{Learning Dynamics.}
Figure~\ref{fig:learning_curves} shows learning curves across all benchmark instances. PPO remains near its initial performance throughout training. Deterministic differentiable planning and JaxPlan exhibit rapid early progress but quickly plateau, indicating that optimization becomes trapped in poor regions of the objective landscape.
Adding stochasticity enables continued improvement. Fixed-noise variants often achieve faster initial gains but may stagnate depending on the noise scale. In contrast, adaptive noise maintains sustained improvement, enabling both faster convergence and continued progress over time.
These effects are more pronounced in harder instances. In \texttt{HVAC20} (Figure~\ref{fig:learning_curves:h5}) and \texttt{Reservoir30} (Figure~\ref{fig:learning_curves:r5}), adaptive noise continues to improve while other methods plateau early. \texttt{PowerGen11} presents a different regime, where relatively small noise is sufficient to find high-quality policies; the adaptive method matches this performance by adjusting the noise level appropriately, as further analyzed below.

\paragraph{Adaptive Noise Analysis.}

\begin{figure*}[t]
    \centering
    \footnotesize
    
    % Row 1: sigma profiles
    \begin{subfigure}{0.32\textwidth}
        \includegraphics[width=\linewidth]{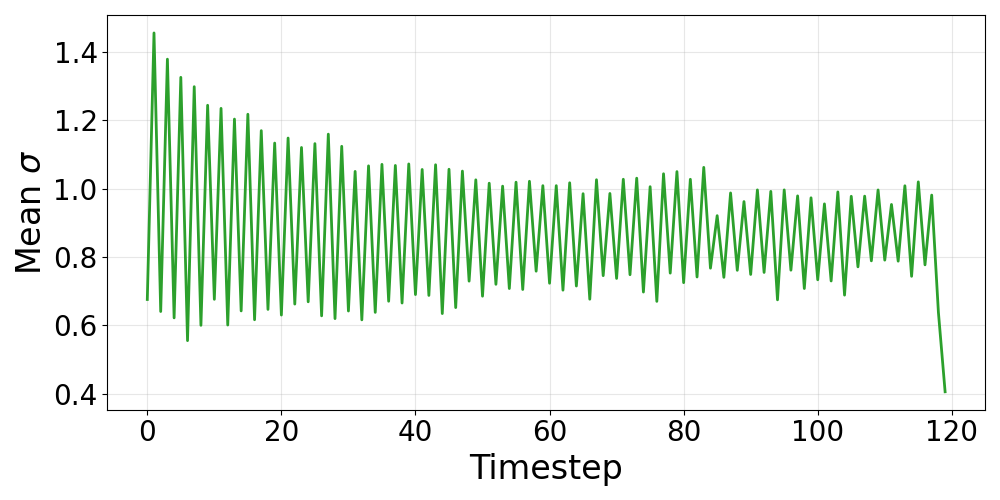}
        % \caption{PowerGen11 Sigma}
        \caption{\texttt{PowerGen11}: average $\sigma_t$  over horizon}
        \label{fig:sigma_heat_prof:ps}
    \end{subfigure}
    \hfill
    \begin{subfigure}{0.32\textwidth}
        \includegraphics[width=\linewidth]{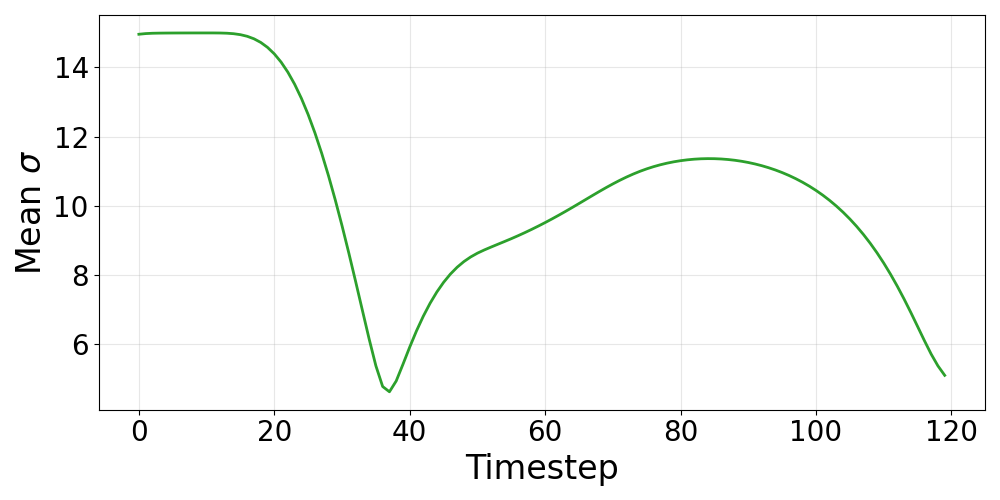}
        % \caption{HVAC20 Sigma}
        \caption{\texttt{HVAC20}: average $\sigma_t$ over \\horizon}
        \label{fig:sigma_heat_prof:hs}
    \end{subfigure}
    \hfill
    \begin{subfigure}{0.32\textwidth}
        \includegraphics[width=\linewidth]{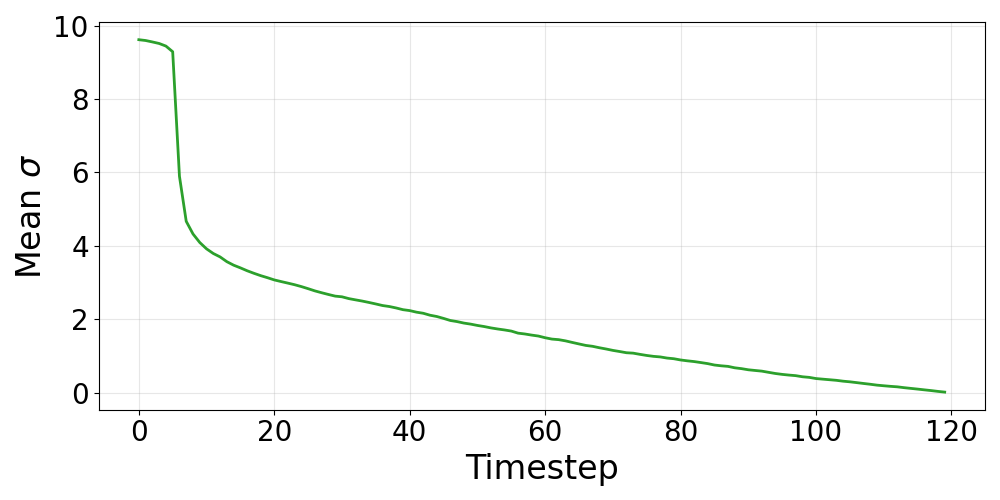}
        % \caption{Reservoir30 Sigma}
        \caption{Reservoir30: average $\sigma_t$ over horizon}
        \label{fig:sigma_heat_prof:rs}
    \end{subfigure}
    
    \vspace{0.5em}
    
    % Row 2: heatmaps
    \begin{subfigure}{0.33\textwidth}
        \includegraphics[width=\linewidth]{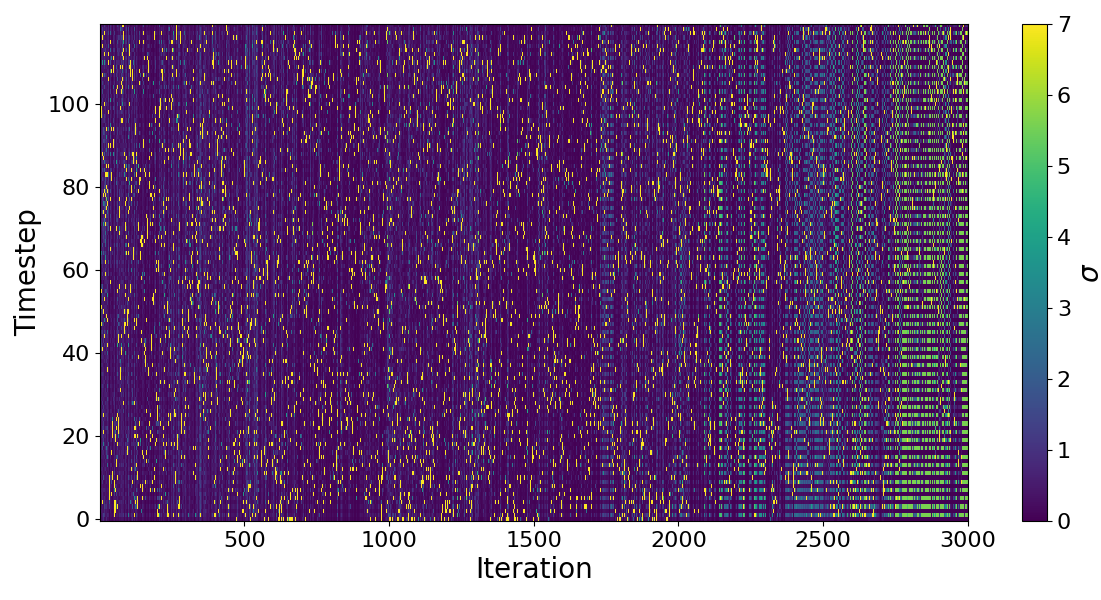}
        \caption{\texttt{PowerGen11}: $\sigma_t$ across \\timesteps and iterations}
        \label{fig:sigma_heat_prof:ph}
    \end{subfigure}
    \hfill
    \begin{subfigure}{0.33\textwidth}
        \includegraphics[width=\linewidth]{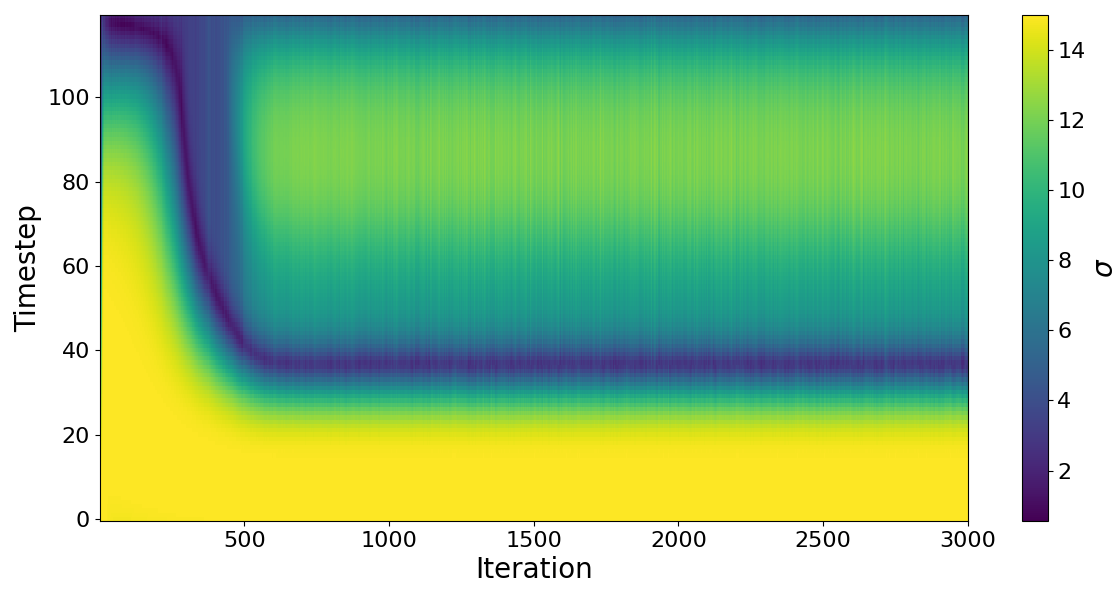}
        \caption{\texttt{HVAC20}: $\sigma_t$ across \\timesteps and iterations}
        \label{fig:sigma_heat_prof:hh}
    \end{subfigure}
    \hfill
    \begin{subfigure}{0.33\textwidth}
        \includegraphics[width=\linewidth]{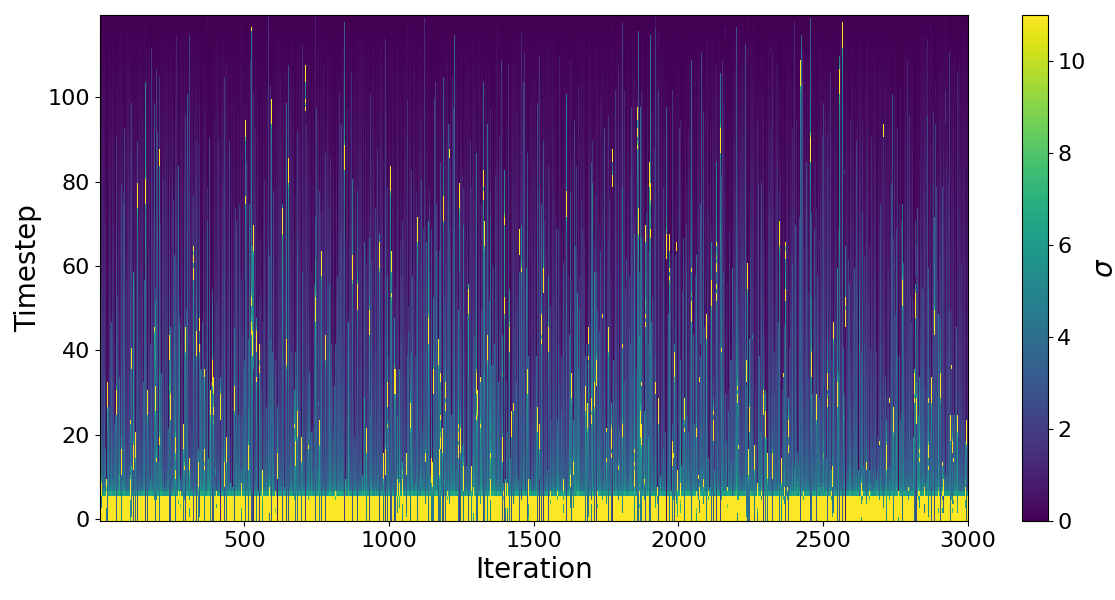}
        \caption{Reservoir30: $\sigma_t$ across \\timesteps and iterations}
        \label{fig:sigma_heat_prof:rh}
    \end{subfigure}
    
    \caption{
    \textbf{Adaptive noise behavior across timesteps and optimization iterations.}
    Top row: average noise magnitude $\sigma_t$ as a function of timestep, averaged over optimization iterations.
    Bottom row: heatmaps showing noise magnitude across timesteps (vertical axis) and optimization iterations (horizontal axis).
    Results are shown for the most challenging instance of each domain.
    The adaptive method produces a structured, time-dependent noise profile that varies across both timesteps and training iterations.
    The allocation of noise is highly dynamic and problem-dependent, with different domains exhibiting distinct patterns.
    }
    
    \label{fig:sigma_heat_prof}
\end{figure*}

To better understand the behavior of the adaptive mechanism, we analyze how the noise magnitude evolves across both timesteps and optimization iterations. We focus on the most challenging instance from each domain.
Figures~\ref{fig:sigma_heat_prof:ps}-\ref{fig:sigma_heat_prof:rs} show the average noise magnitude as a function of the timestep within the planning horizon. 
The results reveal that noise allocation is domain-dependent and reflects the underlying dynamics.
In \texttt{Reservoir30} (Figure~\ref{fig:sigma_heat_prof:rs}), higher noise is concentrated in earlier timesteps, indicating the importance of early decisions.
In \texttt{HVAC20} (Figure~\ref{fig:sigma_heat_prof:hs}), noise is high at early timesteps, decreases sharply over the first third of the horizon, and then follows a non-monotonic variation over the remaining timesteps, suggesting that later decisions intermittently regain importance.
In \texttt{PowerGen11} (Figure~\ref{fig:sigma_heat_prof:ps}), the noise magnitude stabilizes around $\sigma \approx 1$ across the horizon, consistent with the strong performance of fixed noise at this scale (Figure~\ref{fig:learning_curves:pg5}).
Figure~\ref{fig:sigma_heat_prof} 
provides a joint view of how noise is allocated across both timesteps and optimization iterations. Initially, high noise is applied broadly across the trajectory. As optimization progresses, the noise distribution evolves, often becoming more concentrated on subsets of timesteps. This behavior reflects the changing sensitivity of the objective and highlights the dynamic nature of the adaptive mechanism.
These results demonstrate that the adaptive method can automatically capture the structure of the optimization problem, allocating exploration to regions where it appears most beneficial both temporally within the trajectory and across the course of learning.

\section{Conclusion and Discussion}
The empirical results confirm that differentiable planning is fundamentally limited by optimization challenges arising from flat gradients and sharp transitions, particularly in nonlinear and hybrid domains.
By introducing stochastic perturbations in the action space, MDPO mitigates these issues, enabling the optimizer to escape poor local optima and achieve substantially improved performance. The effectiveness of adaptive noise further highlights the importance of tailoring exploration to the local structure of the optimization landscape.
This work demonstrates that the primary limitation of differentiable planning in complex domains lies not only in modeling, but in the optimization dynamics used to solve the resulting problems. Rather than modifying the model or its relaxation, MDPO improves performance by enhancing the optimization process itself through structured, model-driven exploration. This provides a simple and general approach for addressing optimization pathologies in model-based planning.
More broadly, our findings highlight the role of stochastic exploration in differentiable planning. While exploration in this setting differs from its role in reinforcement learning, the parallels suggest a deeper connection between model-based gradient optimization and exploration-driven learning. Bridging these perspectives may offer promising directions for developing more robust and scalable decision-making methods.

{
\small
\bibliographystyle{plain}
\bibliography{bibliography}
}

%%%%%%%%%%%%%%%%%%%%%%%%%%%%%%%%%%%%%%%%%%%%%%%%%%%%%%%%%%%%
\newpage
\appendix

\section{Gradient Derivation for Stochastic Action Perturbations}

In this appendix, we derive the gradient expression stated in Theorem~\ref{thm:grad}.

We consider a stochastic policy obtained by perturbing a deterministic policy with additive noise:
\[
\tilde a_t = \pi_\theta(s_t) + \epsilon_t, 
\quad \epsilon_t \sim \mathcal{N}(0, \Sigma_t),
\]
where the noise is independent of the policy parameters $\theta$. The system evolves according to:
\[
s_{t+1} = f(s_t, \tilde a_t).
\]

The objective is defined as:
\[
J(\theta)
=
\mathbb{E}_{\epsilon}
\left[
\sum_{t=0}^{H-1} r(s_t, \tilde a_t)
\right].
\]

Under standard regularity conditions (e.g., smoothness and boundedness), we interchange gradient and expectation:
\[
\nabla_\theta J(\theta)
=
\mathbb{E}_{\epsilon}
\left[
\sum_{t=0}^{H-1}
\nabla_\theta r(s_t, \tilde a_t)
\right].
\]

Applying the chain rule, each term decomposes as:
\[
\nabla_\theta r(s_t,\tilde a_t)
=
\frac{\partial r}{\partial s_t}\frac{d s_t}{d\theta}
+
\frac{\partial r}{\partial \tilde a_t}\frac{d \tilde a_t}{d\theta},
\]
where $\frac{\partial}{\cdot}$ denotes partial derivatives and $\frac{d}{d\theta}$ denotes total derivatives through the computational graph.

Since $\epsilon_t$ is independent of $\theta$, we have:
\[
\frac{d \tilde a_t}{d\theta}
=
\frac{d \pi_\theta(s_t)}{d\theta}
=
\frac{\partial \pi_\theta(s_t)}{\partial \theta}
+
\frac{\partial \pi_\theta(s_t)}{\partial s_t}\frac{d s_t}{d\theta}.
\]

Differentiating the system dynamics yields:
\[
\frac{d s_{t+1}}{d\theta}
=
\frac{\partial f}{\partial s_t}\frac{d s_t}{d\theta}
+
\frac{\partial f}{\partial \tilde a_t}\frac{d \tilde a_t}{d\theta}.
\]

Substituting the expression for $\frac{d \tilde a_t}{d\theta}$, we obtain:
\[
\frac{d s_{t+1}}{d\theta}
=
\left(
\frac{\partial f}{\partial s_t}
+
\frac{\partial f}{\partial \tilde a_t}
\frac{\partial \pi_\theta}{\partial s_t}
\right)
\frac{d s_t}{d\theta}
+
\frac{\partial f}{\partial \tilde a_t}
\frac{\partial \pi_\theta}{\partial \theta}.
\]

Assuming that the initial state is independent of $\theta$, we have $\frac{d s_0}{d\theta} = 0$. Unrolling the recursion gives:
\[
\frac{d s_t}{d\theta}
=
\sum_{\tau=0}^{t-1}
\left(
\prod_{j=\tau+1}^{t-1}
\left(
\frac{\partial f}{\partial s_j}
+
\frac{\partial f}{\partial \tilde a_j}
\frac{\partial \pi_\theta}{\partial s_j}
\right)
\right)
\frac{\partial f}{\partial \tilde a_\tau}
\frac{\partial \pi_\theta}{\partial \theta},
\]
with the convention that an empty product equals the identity matrix.

\paragraph{Connection to Theorem~\ref{thm:grad}.}
Substituting this expression into the gradient expansion, we obtain:
\[
\nabla_\theta J(\theta)
=
\mathbb{E}_{\epsilon}
\left[
\sum_{t=0}^{H-1}
\left(
\frac{\partial r}{\partial s_t}\frac{d s_t}{d\theta}
+
\frac{\partial r}{\partial \tilde a_t}\frac{d \tilde a_t}{d\theta}
\right)
\right].
\]

The first term captures the \emph{indirect effect} of $\theta$ through the state trajectory:
\[
\frac{\partial r}{\partial s_t}\frac{d s_t}{d\theta},
\]
which, after substitution, corresponds to the accumulation of sensitivity through the dynamics.

The second term captures the \emph{direct effect} through the action:
\[
\frac{\partial r}{\partial \tilde a_t}\frac{d \tilde a_t}{d\theta}
=
\frac{\partial r}{\partial \tilde a_t}
\left(
\frac{\partial \pi_\theta}{\partial \theta}
+
\frac{\partial \pi_\theta}{\partial s_t}\frac{d s_t}{d\theta}
\right).
\]

Combining these contributions yields the expression stated in Theorem~\ref{thm:grad}, consisting of:
(i) a direct term corresponding to the immediate influence of actions on rewards, and  
(ii) an indirect term capturing the propagation of parameter influence through the system dynamics.

This establishes that the gradient can be computed via standard backpropagation through the stochastic computation graph under reparameterized action noise.

\section{Noise Extension to Multiple Action Types}

In domains such as \texttt{HVAC}, the action at time \(t\) is not a single vector but a collection of action tensors. For example, one may have
\[
a_t = \bigl(a_t^{(1)}, a_t^{(2)}, \dots, a_t^{(M)}\bigr),
\]
where each \(a_t^{(m)}\), for \(m=1,\dots,M\), corresponds to a different action fluent type. In \texttt{HVAC}, these are the action tensors associated with \texttt{fan-in(zone)} and \texttt{heat-input(heater)}.
which can be viewed as a structured matrix (or collection of tensors), where each block \(a_t^{(m)}\) corresponds to a different action fluent type.
\\\\ 
The trajectory objective remains a scalar,
\[
J(\tau),
\]
but we now differentiate it separately with respect to each action tensor:
\[
g_t^{(m)} = \nabla_{a_t^{(m)}} J(\tau),
\qquad m=1,\dots,M.
\]

Each such gradient is converted into a scalar action-level sensitivity:
\[
c_t^{(m)} = \left\| g_t^{(m)} \right\|_2.
\]

These action-level sensitivities are then aggregated into a single timestep score. In the current implementation, the default aggregation is the mean:
\[
s_t = \frac{1}{M}\sum_{m=1}^{M} c_t^{(m)}
=
\frac{1}{M}\sum_{m=1}^{M} \left\| \nabla_{a_t^{(m)}} J(\tau) \right\|_2.
\]

Once \(s_t\) is obtained, the rest of the pipeline is unchanged: the timestep scores are normalized across the trajectory, mapped to a scalar noise level \(\sigma_t\), and this same \(\sigma_t\) is then broadcast to all action tensors at timestep \(t\). Thus, in the multi-action setting, the current method still produces one scalar noise level per timestep, not one separate noise level per action type.

\section{Benchmark Domains}
\subsection{Power Generation Domain}

The Power Generation (PowerGen) domain is a continuous relaxation of the classical unit commitment problem, where a set of generators must meet a stochastic demand while minimizing operational costs. The system evolves over a finite horizon $H$, with dynamics defined over continuous and discrete state variables.

\paragraph{State and action space.}
Let $\mathcal{P}$ denote the set of generators. For each generator $p \in \mathcal{P}$, the system maintains:
\begin{itemize}
    \item[-] a continuous production level from the previous timestep, $x^{(p)}_t \in \mathbb{R}_{\geq 0}$,
    \item[-] a binary activation state $o^{(p)}_t \in \{0,1\}$.
\end{itemize}
% - a continuous production level from the previous timestep, $x^{(p)}_t \in \mathbb{R}_{\geq 0}$,
% - a binary activation state $o^{(p)}_t \in \{0,1\}$.

Additionally, the system includes a continuous exogenous temperature variable $T_t \in \mathbb{R}$.

At each timestep, the controller selects a continuous action:
\[
a^{(p)}_t \in [0, a^{(p)}_{\max}],
\]
representing the intended production level of generator $p$.

\paragraph{Activation and effective production.}
A generator is active if its production exceeds a minimum threshold:
\[
o^{(p)}_t = \mathbb{I}\left[a^{(p)}_t \geq a^{(p)}_{\min}\right].
\]

The effective production is defined as:
\[
\tilde{a}^{(p)}_t =
\begin{cases}
\mathrm{clip}\left(a^{(p)}_t + \eta^{(p)}_t, \; a^{(p)}_{\min}, \; a^{(p)}_{\max} \right), & \text{if } o^{(p)}_t = 1, \\
0, & \text{otherwise},
\end{cases}
\]
where $\eta^{(p)}_t \sim \text{Weibull}(k_p, \lambda_p)$ models stochastic fluctuations in production.

\paragraph{State transitions.}
The production and activation states evolve as:
\[
x^{(p)}_{t+1} = \tilde{a}^{(p)}_t, \quad
o^{(p)}_{t+1} = o^{(p)}_t.
\]

The temperature evolves stochastically:
\[
T_{t+1} = \mathrm{clip}\big( T_t + \xi_t, \; T_{\min}, \; T_{\max} \big), 
\quad \xi_t \sim \mathcal{N}(0, \sigma_T^2).
\]

\paragraph{Demand model.}
Demand is a nonlinear function of temperature:
\[
D_t = D_0 + \alpha (T_t - T^*)^2,
\]
where $T^*$ is the reference temperature minimizing demand.

The fulfilled demand is given by:
\[
F_t = \min\left( D_t, \sum_{p \in \mathcal{P}} \tilde{a}^{(p)}_t \right).
\]

\paragraph{Reward function.}
The reward at each timestep combines multiple components:

\[
R_t =
- \sum_{p \in \mathcal{P}} c_p \tilde{a}^{(p)}_t
+ \rho F_t
- \mathbb{I}[D_t > F_t] \cdot P
- \sum_{p \in \mathcal{P}} \lambda_p \left| \tilde{a}^{(p)}_t - x^{(p)}_t \right|
- \sum_{p \in \mathcal{P}} \kappa_p \mathbb{I}\left[(1 - o^{(p)}_t) \cdot o^{(p)}_{t+1}\right],
\]
where:
\begin{itemize}
    \item[-] $c_p$ is the production cost per unit,
    \item[-] $\rho$ is the revenue per unit of fulfilled demand,
    \item[-] $P$ is the penalty for unmet demand,
    \item[-] $\lambda_p$ penalizes changes in production,
    \item[-] $\kappa_p$ penalizes switching a generator on.
\end{itemize}
% - $c_p$ is the production cost per unit,
% - $\rho$ is the revenue per unit of fulfilled demand,
% - $P$ is the penalty for unmet demand,
% - $\lambda_p$ penalizes changes in production,
% - $\kappa_p$ penalizes switching a generator on.

\subsection{HVAC Domain}

The \texttt{HVAC} domain models the control of temperature in a network of interconnected zones subject to energy costs and comfort constraints. The system evolves over a finite horizon $H$, with nonlinear dynamics and stochastic occupancy.

\paragraph{State and action space.}
Let $\mathcal{Z}$ denote the set of zones and $\mathcal{H}$ the set of heaters. The state consists of:
\begin{itemize}
    \item[-] zone temperatures $\delta^{(z)}_t \in \mathbb{R}$ for each $z \in \mathcal{Z}$,
    \item[-] heater temperatures $\theta^{(h)}_t \in \mathbb{R}$ for each $h \in \mathcal{H}$,
    \item[-] occupancy variables $o^{(z)}_t \in \{0,1\}$ indicating whether zone $z$ is occupied.
\end{itemize}
% - zone temperatures $\delta^{(z)}_t \in \mathbb{R}$ for each $z \in \mathcal{Z}$,
% - heater temperatures $\theta^{(h)}_t \in \mathbb{R}$ for each $h \in \mathcal{H}$,
% - occupancy variables $o^{(z)}_t \in \{0,1\}$ indicating whether zone $z$ is occupied.

The control actions are continuous:
\begin{itemize}
    \item[-] airflow into each zone $a^{(z)}_t \geq a_{\min}$,
    \item[-] heat input to each heater $u^{(h)}_t \in \mathbb{R}$.
\end{itemize}
% - airflow into each zone $a^{(z)}_t \geq a_{\min}$,
% - heat input to each heater $u^{(h)}_t \in \mathbb{R}$.

\paragraph{Occupancy dynamics.}
Occupancy evolves stochastically according to:
\[
o^{(z)}_{t+1} = o^{(z)}_t \cdot \xi^{(z)}_t, 
\quad \xi^{(z)}_t \sim \text{Bernoulli}(1 - p^{(z)}),
\]
where $p^{(z)}$ is the probability of becoming unoccupied.

\paragraph{Heater dynamics.}
Each heater aggregates incoming air from connected zones. Let $\mathcal{Z}(h)$ denote the zones connected to heater $h$. The heater temperature evolves as:
\[
\theta^{(h)}_{t+1} =
- \kappa_h (\theta^{(h)}_t)^2
+ \theta^{(h)}_t
+ \frac{\Delta t}{V_h} \bar{a}^{(h)}_t \left( \bar{\delta}^{(h)}_t - T_{\text{out}} \right)
+ \frac{\Delta t}{K V_h} \tilde{u}^{(h)}_t,
\]
where:
\begin{itemize}
    \item[-] $\bar{a}^{(h)}_t$ is the average airflow into heater $h$,
    \item[-] $\bar{\delta}^{(h)}_t$ is the average temperature of connected zones,
    \item[-] $\tilde{u}^{(h)}_t$ is a clipped version of the heat input,
    \item[-] $\kappa_h$ models heat dissipation.
\end{itemize}
% - $\bar{a}^{(h)}_t$ is the average airflow into heater $h$,
% - $\bar{\delta}^{(h)}_t$ is the average temperature of connected zones,
% - $\tilde{u}^{(h)}_t$ is a clipped version of the heat input,
% - $\kappa_h$ models heat dissipation.

\paragraph{Zone dynamics.}
Zone temperatures evolve according to nonlinear heat transfer:
\[
\delta^{(z)}_{t+1} =
- \kappa_z (\delta^{(z)}_t - T_{\text{out}})^2
+ \delta^{(z)}_t
+ \frac{\Delta t}{V_z} a^{(z)}_t \sum_{h \in \mathcal{H}(z)} 
\frac{\theta^{(h)}_t - \delta^{(z)}_t}{|\mathcal{H}(z)|}
+ \Delta t \sum_{z' \in \mathcal{Z}} 
\gamma_{z,z'} (\delta^{(z')}_t - \delta^{(z)}_t),
\]
where:
\begin{itemize}
    \item[-] $\mathcal{H}(z)$ denotes heaters connected to zone $z$,
    \item[-] $\gamma_{z,z'}$ represents thermal conductivity between adjacent zones,
    \item[-] $V_z$ is the zone volume.
\end{itemize}
% - $\mathcal{H}(z)$ denotes heaters connected to zone $z$,
% - $\gamma_{z,z'}$ represents thermal conductivity between adjacent zones,
% - $V_z$ is the zone volume.

\paragraph{Reward function.}
The reward penalizes energy consumption and temperature deviations from a comfort range $[\delta^{(z)}_{\min}, \delta^{(z)}_{\max}]$:
\[
R_t =
- \sum_{h \in \mathcal{H}} c_h \, (\tilde{u}^{(h)}_t)^2
- c_a \sum_{z \in \mathcal{Z}} (a^{(z)}_t)^2
- \lambda \sum_{z \in \mathcal{Z}} \mathbb{I}[o^{(z)}_t = 1] \cdot \phi(\delta^{(z)}_t),
\]
where the discomfort penalty is:
\[
\phi(\delta) =
\begin{cases}
(\delta - \delta_{\min})^2, & \delta < \delta_{\min}, \\
(\delta - \delta_{\max})^2, & \delta > \delta_{\max}, \\
0, & \text{otherwise}.
\end{cases}
\]

\subsection{Reservoir Control Domain}

The \texttt{Reservoir} Control domain models the regulation of water levels in a network of interconnected reservoirs subject to stochastic inflow, evaporation, and physical constraints. The system evolves over a finite horizon $H$.

\paragraph{State and action space.}
Let $\mathcal{R}$ denote the set of reservoirs. The state consists of continuous water levels
\[
\ell^{(r)}_t \in [0, \ell^{(r)}_{\max}], \quad r \in \mathcal{R}.
\]

At each timestep, the controller selects a continuous action
\[
a^{(r)}_t \in [0, \ell^{(r)}_{\max}],
\]
representing the amount of water released from reservoir $r$.

\paragraph{Stochastic inflow and evaporation.}
Each reservoir receives stochastic rainfall:
\[
r^{(r)}_t = | \xi^{(r)}_t |, \quad \xi^{(r)}_t \sim \mathcal{N}(0, \sigma_r^2),
\]
and loses water due to evaporation:
\[
e^{(r)}_t = \beta \cdot \frac{\ell^{(r)}_t}{\ell^{(r)}_{\text{top}}},
\]
where $\beta$ is the evaporation factor.

\paragraph{Flow and constraints.}
The effective release is clipped to feasible bounds:
\[
\tilde{a}^{(r)}_t = \min\left(\ell^{(r)}_t, \max(0, a^{(r)}_t)\right).
\]

Water is distributed across downstream reservoirs according to the connectivity graph. Let $\mathcal{D}(r)$ denote the set of reservoirs downstream of $r$. The flow from $r$ to each downstream reservoir is:
\[
f^{(r)}_t = \frac{\tilde{a}^{(r)}_t}{|\mathcal{D}(r)| + \mathbb{I}[\text{sea}(r)]}.
\]

The total inflow to reservoir $r$ is:
\[
\text{inflow}^{(r)}_t = \sum_{r' \in \mathcal{U}(r)} f^{(r')}_t,
\]
where $\mathcal{U}(r)$ denotes upstream reservoirs.

Overflow is defined as:
\[
\text{overflow}^{(r)}_t = \max\left(0, \ell^{(r)}_t - \tilde{a}^{(r)}_t - \ell^{(r)}_{\text{top}} \right).
\]

\paragraph{State transition.}
The reservoir level evolves as:
\[
\ell^{(r)}_{t+1} =
\min\left(\ell^{(r)}_{\text{top}}, 
\max\left(0, \ell^{(r)}_t + \text{inflow}^{(r)}_t + r^{(r)}_t - e^{(r)}_t - \tilde{a}^{(r)}_t - \text{overflow}^{(r)}_t \right)\right).
\]

\paragraph{Reward function.}
The objective is to maintain reservoir levels within a desired range $[\ell^{(r)}_{\min}, \ell^{(r)}_{\max}]$. The reward penalizes deviations using piecewise linear costs:
\[
R_t = \sum_{r \in \mathcal{R}} \psi(\ell^{(r)}_{t+1}),
\]
where
\[
\psi(\ell) =
\begin{cases}
c^{(r)}_{\text{low}} \cdot (\ell^{(r)}_{\min} - \ell), & \ell < \ell^{(r)}_{\min}, \\
c^{(r)}_{\text{high}} \cdot (\ell - \ell^{(r)}_{\max}), & \ell^{(r)}_{\max} < \ell \leq \ell^{(r)}_{\text{top}}, \\
c^{(r)}_{\text{high}} \cdot (\ell - \ell^{(r)}_{\max}) + c^{(r)}_{\text{overflow}} \cdot \text{overflow}, & \ell > \ell^{(r)}_{\text{top}}, \\
0, & \text{otherwise}.
\end{cases}
\]

%%%%%%%%%%%%%%%%%%%%%%%%%%%%%%%%%%%%%%%%%%%%%%%%%%%%%%%%%%%%

%\newpage
%\input{checklist.tex}

\end{document}